\documentclass{article}

\usepackage{amsmath,amsfonts,bm}

\def\eqref#1{equation~\ref{#1}}
\def\1{\bm{1}}

\DeclareMathAlphabet{\mathsfit}{\encodingdefault}{\sfdefault}{m}{sl}
\SetMathAlphabet{\mathsfit}{bold}{\encodingdefault}{\sfdefault}{bx}{n}

\DeclareMathOperator*{\argmin}{arg\,min}

\usepackage{adjustbox}
\usepackage[most]{tcolorbox}
\usepackage{arydshln}
\usepackage{amssymb}
\usepackage{tabularx}
\usepackage{enumitem}
\usepackage{graphicx}
\usepackage{colortbl}
\usepackage{multirow}
\usepackage{svg}
\usepackage{MnSymbol}
\usepackage{amsmath}
\usepackage{subcaption}
\usepackage{algorithmicx}
\usepackage{algcompatible}
\usepackage{algorithm}
\usepackage{amssymb}
\usepackage{bbm}
\usepackage[utf8]{inputenc} % allow utf-8 input
\usepackage[T1]{fontenc}    % use 8-bit T1 fonts
\usepackage{booktabs}       % professional-quality tables
\usepackage{amsfonts}       % blackboard math symbols
\usepackage{nicefrac}       % compact symbols for 1/2, etc.
\usepackage{microtype}      % microtypography
\usepackage{xcolor}         % colors

 \usepackage[main, final]{neurips_2026}
\usepackage[utf8]{inputenc} % allow utf-8 input
\usepackage[T1]{fontenc}    % use 8-bit T1 fonts
\usepackage{booktabs}       % professional-quality tables
\usepackage{amsfonts}       % blackboard math symbols
\usepackage{nicefrac}       % compact symbols for 1/2, etc.
\usepackage{microtype}      % microtypography
\usepackage{xcolor}         % colors

\title{Towards Fairness under Label Bias in Image Segmentation: Impact, Measurement and Mitigation}

\author{%
  Aditya Parikh\quad Stella Frank\quad Sneha Das\quad Aasa Feragen \\
  Department of Applied Mathematics and Computer Science \\
  Technical University of Denmark \\
  \texttt{adipa@dtu.dk}
}

\begin{document}

\maketitle

\begin{abstract}

Biases in annotation pipelines may introduce \textit{label bias}: when annotation quality or style varies systematically across demographic subgroups, causing disparities in performance and auditability. Label bias in image segmentation remains underexplored, as even detecting it typically requires clean, unbiased annotations, which are not readily available. We present a Confident Learning framework adapted to segmentation for auditing label bias directly in the training data without a clean, unbiased ground truth. By comparing the provided training labels to the model's confident predictions, we measure the prevalence and direction of suspected label errors; where standard overlap metrics like Dice fail. We further show that label bias influences subgroup separability in the encoder's feature space, an artifact we leverage for bias mitigation rather than suppressing it. We evaluate three datasets spanning from synthetic to real-life bias and across experimental conditions, showing how our framework produces proxy audit signals without clean labels and mitigates disparities with a correctly identified cleaner reference group. 

To ensure reproducibility, all code, dataset configuration, and random seed made available \footnote{Available at: https://github.com/ADE-17/label-bias-segmentation}.
\end{abstract}

\section{Introduction}
\label{sec:intro}

Annotation (interchangeably, labels) quality depends on task objective, annotation guidelines, rater experience, labeling tools, and automated initialization procedures \citep{zepf2023label,nichyporuk2022rethinking,zhang2020disentangling}. When these factors are systematically associated with \textit{sensitive attributes} (like race, sex, or age~\citep{sensitive_attributes}), the resulting dataset exhibits \textbf{group-conditioned label bias}. Unlike label noise, label bias is systematic: its prevalence or direction depends on the subgroup or annotation process \citep{wang2021fair,beck2020representation}.

Label bias in safety-critical AI systems can cause substantial harm to affected groups, but identifying it is difficult without access to correct labels. Prior work in medical image segmentation shows that if a model systematically over-segments tumors for certain patient populations, those errors propagate directly into clinical decisions like surgery or radiotherapy planning, harming the affected group~\citep{chen2023algorithmic,koccak2025bias}. Yet, standard segmentation metrics (e.g., Dice, IoU) condense complex errors into a single scalar, masking where and how bias manifests~\citep{taha2015metrics,reinke2021common}. It becomes more difficult to detect label bias when the data lack clean, unbiased ground-truth labels, which leads to systematic bias~\citep{quinn2023interobserver,parikh2025investigating,juni2026false}.

Label bias is well understood in classification~\citep{obermeyer2019dissecting}, where controlled label-flip simulations~\citep{dai2020label} and detection tools such as Cleanlab~\citep{northcutt2021confident,lad2023estimating} establish a principled experimental foundation~\citep{zhang2024mitigating,wang2025toward}. In segmentation, errors are spatially correlated and often concentrated at object boundaries, rather than occurring as independent label flips \citep{heller2018imperfect,yao2023spatial}, creating corruption that is more challenging than spatially unstructured noise \citep{heller2018imperfect}. A research gap exists for simulating, identifying, and mitigating group-conditioned label bias in segmentation.

We address this gap with a framework to audit and mitigate group-conditional label bias. \textbf{Our contributions} are:

\quad (i) We adapt Confident Learning to segmentation as a fairness auditing tool, introducing metrics to estimate the magnitude and spatial direction of potential label errors \textbf{without clean ground-truth labels}.

\quad (ii) We show that label bias amplifies subgroup separability in segmentation, and its effect on learned models and fairness.

\quad (iii) We propose a subgroup-conditioned segmentation approach that leverages a reference annotation style to mitigate label bias, and present a comparison against popular baselines across various experimental settings.

\section{Background and Related Work}
\label{sec:related}

\paragraph{Label bias.} Label bias appears when the labels used for training and testing are systematically incorrect in ways that differ across sensitive groups. Existing work highlights how label bias will affect both model performance~\citep{jiang2020identifying,vorontsov2021label,dai2020label} and methods for auditing and explaining AI models~\citep{parikh2025investigating, akintande2025ewaf}, but existing mitigation tools are limited to settings where true labels are available~\citep{parikh2025investigating} or the bias is known~\citep{jiang2020identifying}. As label bias primarily affects settings where true labels are impossible or expensive to obtain, this leaves a crucial unmet need for mitigation when true labels or bias are unknown. 

\paragraph{Learning with Noisy Labels.} While label noise covers generic disagreement between observed and intended labels, label bias refers to systematic label errors associated with a specific group, annotator, tool, or protocol~\citep{dai2020label,wang2021fair,beck2020representation}. This systematic corruption has fundamental consequences for learning: while models are robust to label noise, they are sensitive to label bias even at modest scales~\citep{vorontsov2021label}. The field of learning with noisy labels (LNL) has focused on training robust neural networks under corrupted annotations. LNL traditionally models noise as uniform random flipping or class-conditional label transitions~\citep{natarajan2013learning, patrini2017making}; for segmentation, it is commonly treated as boundary uncertainty or random spatial perturbations~\citep{bernhard2024s}. Prior work has highlighted that annotation errors in segmentation are rarely random~\citep{lipton2018mythos}, making standard LNL techniques, including robust loss functions and global label smoothing, fail, as they assume noise is independent of underlying semantic content~\citep{jiang2020beyond,northcutt2021confident}. Our work formalizes this systematic label bias not as a random dataset nuisance, but as a group-conditional geometric bias that alters the network's representation space.

\paragraph{Algorithmic Fairness in Segmentation.} Algorithmic fairness has been extensively studied in classification, with a predominant focus on mitigating performance disparities across protected attributes~\citep{barocas2023fairness,buolamwini2018gender}. Fairness in segmentation remains underexplored~\citep{wang2025toward}. The existing segmentation fairness literature assumes a clean, unbiased ground truth, treating performance disparities as a failure of model capacity alone~\citep{oakden2020hidden}; an assumption we directly challenge. In medical imaging and real-world vision tasks, annotation protocols are inherently subjective and susceptible to demographic-specific boundary distortions~\citep{jungo2018effect,quinn2023interobserver}, such that optimizing blindly on aggregated metrics like Dice under biased labels actively forces the model to encode demographic prejudice~\citep{liu2024we}. By introducing directional metrics to detect label bias, our framework reorients the focus from model-centric fairness toward data-centric bias auditing.

\paragraph{Confident Learning \& Dataset Auditing.} Confident Learning (CL) is a dataset auditing framework~\citep{northcutt2021confident} utilizing the joint distribution over noisy and true labels. CL has, e.g., been applied to highlight annotation errors for review~\citep{lad2023estimating}, and also for medical image segmentation~\citep{zhang2020characterizing,pethmunee2025confident}, but most work considers uniform quality issues targeting general robustness. An exception is \citet{li2023mitigating}, which uses CL as a bias mitigation tool for classification. Here, we utilize CL to solve a far more challenging problem: Diagnosing 
label bias in image segmentation, without knowing the nature of our underlying label bias, by projecting the joint distribution matrix calculated by CL onto directional axes to isolate subgroup-conditional bias in segmentation.

\section{Problem Setup and Bias Simulation}
\label{sec:formalizing}

We focus on \textit{subgroup-conditional label bias} -- hereafter \textit{label bias}, where systematic discrepancies between observed and intended annotations occur disproportionately across demographic subgroups.

\paragraph{Notation and Definitions.} Let data \(\mathcal{D}=\{(x_i,y_{\mathrm{obs},i},g_i)\}_{i=1}^{N}\), where \(x_i\in\mathbb{R}^{H\times W\times3}\), \(y_{\mathrm{obs},i}\in\{0,1\}^{H\times W}\) is the observed annotation, and \(g_i\in\mathcal{G}=\{g_c,g_b\}\) denotes the selected reference and potentially biased groups. The latent intended mask is denoted by \(y_{\mathrm{true},i}\), with foreground set \(Y_{\mathrm{true},i}\). It is available only in controlled experiments and is not used by the proposed audit or mitigation procedure. Pixels are indexed by \(p\in\Omega=\{1,\ldots,H\}\times\{1,\ldots,W\}\), and \(a\in\{0,1\}\) indexes the background and foreground classes. For \(p\in\Omega\), \(y_{\mathrm{obs},i,p}\) denotes the pixel label, and
\(Y_{\mathrm{obs},i}=\{p\in\Omega:y_{\mathrm{obs},i,p}=1\}\) its foreground set. Feature channels are indexed by \(k\), and \(\mathbf{1}[\cdot]\) denotes the indicator function.

\paragraph{Sample-level Label Bias.} The bias ratio $\beta\in[0,1]$ denotes the proportion of samples in \(g_b\) whose entire masks are corrupted, rather than the proportion of corrupted pixels (i.e, at sample level, maintaining consistency with label bias definitions in classification fairness literature~\citep{jorgensen2026detecting}).

\paragraph{Simulating Morphological Bias.} We draw on evidence from medical image analysis, where the most prevalent form of label bias is systematic under- or over-segmentation of anatomical structures~\citep{benvcevic2024understanding, pesce2020interobserver}. When annotators encounter images with lower visual contrasts, they frequently compensate by drawing overly conservative or overly expansive boundaries, e.g.~in specific skin tones in dermatology~\citep{benvcevic2024understanding}, varying tissue densities in radiology~\citep{pesce2020interobserver}, or delineating images under the influence of external conditions~\citep{weltens2001interobserver}, e.g., low lighting, geographic regions~\citep{schumann2023consensus}.

In controlled experiments, we replicate: (i)~\textit{dilation} that expands boundaries, simulating over-annotation; (ii)~\textit{erosion} that contracts boundaries, simulating under-annotation; (iii)~\textit{HBD} produces spatially varying mixed over- and under-annotation, simulating observed non-uniform annotation inconsistencies (details in Appendix~\ref{appendix:hbd}). All morphological operations are parameterized by a radius $r_d$, applied to $g_b$ at a sample-level bias rate $\beta$ to reproduce an approximation of real-world label bias whose severity can be systematically varied.

% we draw \(z_i\sim\operatorname{Bernoulli}(\beta)\) and construct 
% \begin{equation}
% y_{\mathrm{obs},i} = \begin{cases} T_{r_d}(y_{\mathrm{true},i}), & g_i=g_b \text{ and } z_i=1,\\ y_{\mathrm{true},i}, & \text{otherwise}, \end{cases} \label{eq:bias_injection} 
% \end{equation}
% where \(T_{r_d}\) denotes erosion, dilation,~\citep{soille2004erosion} or harmonic boundary deformation with radius \(r_d\). Thus, \(\beta\) controls bias prevalence, while \(T_{r_d}\) controls its geometric direction and severity, applied to $g_b$. \textit{Dilation} expands boundaries, simulating over-annotation; \textit{Erosion} contracts boundaries, simulating under-annotation; \textit{HBD} produces spatially varying mixed over- and under-annotation, simulating non-uniform annotation inconsistencies observed (details in Appendix~\ref{appendix:hbd}).

\section{A Framework for Auditing and Mitigating Label Bias in Segmentation}
\label{sec:framework}

The proposed framework addresses two stages: \textit{auditing} label bias directly from the training data, and \textit{mitigation}, which corrects for the detected bias during model training. Neither stage accesses \(y_{\mathrm{true}}\) during operation; mitigation
additionally assumes that a reference annotation style can be selected.

\subsection{Auditing Label Bias via Confident Learning}
\label{sec:cl_audit}

\begin{figure}[t]
    \centering
    \includegraphics[width=\linewidth]{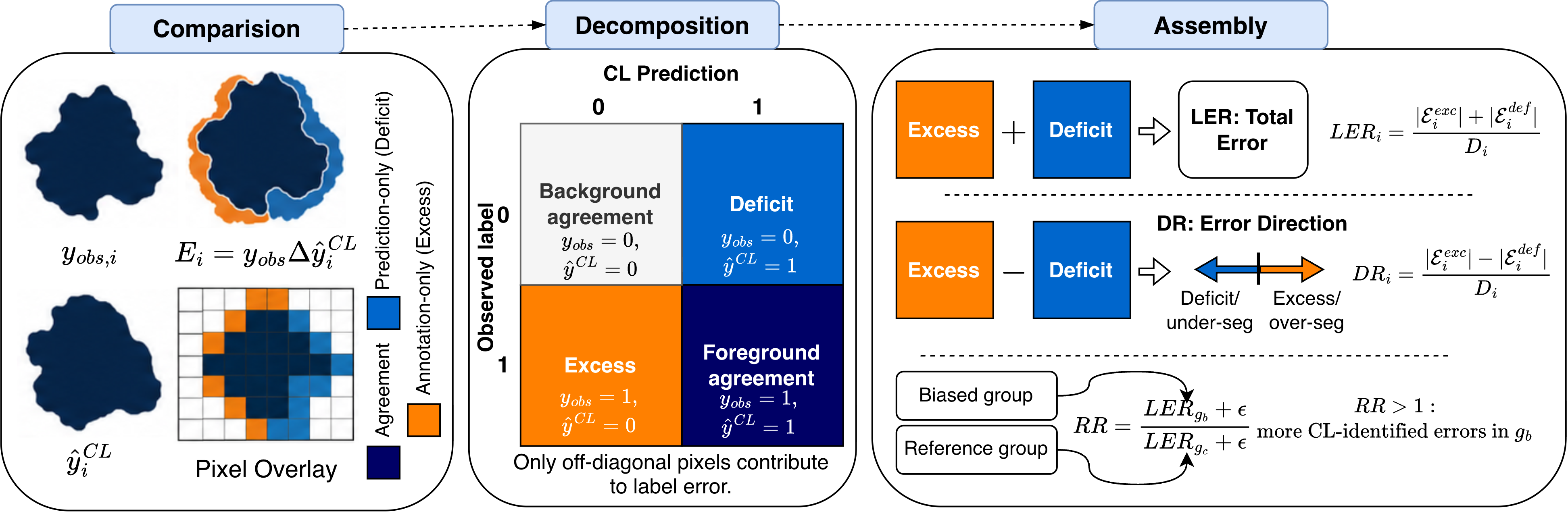}
    \caption{
    \textbf{CL-based decomposition of annotation disagreement.}
Observed labels \(y_{\mathrm{obs}}\) are compared with confident assignments
derived from cross-fitted model predictions.
}
    \label{fig:cl_decomposition}
    % \vspace{-6pt}
\end{figure}

We adapt Confident Learning (CL)~\citep{northcutt2021confident} to audit group-conditional annotation disagreement in segmentation. 
The idea behind CL is to identify data points (pixels) for which model predictions $\hat{y}_{i,p}$ \emph{confidently} disagree with the annotated labels $y_{obs,i,p}$. To this end, model predictive probabilities \(\widehat p_{i,p}(a)\) are estimated using a \(K\)-fold cross-fitting procedure.
Confident predictions are those over the class-specific confidence threshold, corresponding to the empirical mean probability of that class for points belonging to that class in 
\begin{equation}
P_a = \{(i,p): y_{\mathrm{obs},i,p}=a\} \quad
\widehat\tau_a = \frac{1}{|\mathcal P_a|} \sum_{(i,p)\in\mathcal P_a} \widehat p_{i,p}(a).
\end{equation}
Each pixel is then either assigned to the class with the highest model probability that also exceeds the class confidence threshold, or falls back to $\arg\max$ if no probability exceeds the threshold:
\begin{equation}
\widehat y^{\mathrm{CL}}_{i,p} = \begin{cases}
\underset{a\in\{0,1\}:\, \widehat p_{i,p}(a)\geq\widehat{\tau}_a}{\arg\max}\; \widehat p_{i,p}(a), & \text{if } \max_{a} \left(\widehat p_{i,p}(a) - \widehat\tau_a\right) \geq 0, \\[6pt]
\underset{a\in\{0,1\}}{\arg\max}\; \widehat p_{i,p}(a), & \text{otherwise}.
\end{cases}
\end{equation}
We can then define a ``confident error set'' made up of those pixels in an image with confident prediction-annotation disagreements: 
\begin{equation}
\mathcal E_i = \left\{ p\in\Omega: \widehat p_{i,p}(\widehat y^{\mathrm{CL}}_{i,p})\geq\widehat{\tau}_{\widehat y^{\mathrm{CL}}_{i,p}}\ \text{and}\ y_{\mathrm{obs},i,p}\neq\widehat y^{\mathrm{CL}}_{i,p} \right\},\quad 
\widehat Y_i^{\mathrm{CL}} = \left\{ p\in\Omega: \widehat y^{\mathrm{CL}}_{i,p}=1 \right\}.
\label{eq:cl_error_set}
\end{equation}

%For each \((x_i,y_{\mathrm{obs},i},g_i)\), \(K\)-fold cross-fitting produces pixel-wise probabilities \(\widehat p_{i,p}(a)\), \(a\in\{0,1\}\), using a model that was not trained on image \(i\). CL uses these out-of-sample probabilities to identify confident disagreements with \(y_{\mathrm{obs},i}\). A $\hat{}$ (hat) indicates a CL-based proxy estimate rather than a true quantity by using \(y_{\mathrm{true}}\); hats are omitted in tables.

%\paragraph{Confident Class Assignments.} For each class \(a\in\{0,1\}\), CL estimates a confidence threshold
%\begin{equation} 
%\widehat\tau_a = \frac{1}{|\mathcal P_a|} \sum_{(i,p)\in\mathcal P_a} \widehat p_{i,p}(a), \quad \mathcal P_a = \{(i,p):y_{\mathrm{obs},i,p}=a\}. \label{eq:cl_threshold} 
%\end{equation} 
%A pixel is confidently assigned to the class with the highest probability:
%\begin{equation}
%   \widehat y^{\mathrm{CL}}_{i,p} = \underset{ a\in\{0,1\}:\, \widehat p_{i,p}(a)\geq\widehat{\tau}_a }{\arg\max}\; \widehat p_{i,p}(a). \label{eq:cl_assignment} 
%   \end{equation}
%   A pixel is excluded from the potential error set if no class meets its threshold. We then define:
%   \begin{equation}
%   \mathcal E_i = \left\{ p\in\Omega: \widehat y^{\mathrm{CL}}_{i,p}\ \text{is defined and}\ y_{\mathrm{obs},i,p}\neq\widehat y^{\mathrm{CL}}_{i,p} \right\},\quad 
%   \widehat Y_i^{\mathrm{CL}} = \left\{ p\in\Omega: \widehat y^{\mathrm{CL}}_{i,p}=1 \right\}.
%   \label{eq:cl_error_set}
%   \end{equation}
%   Thus, \(\mathcal E_i\) contains only confident prediction-annotation disagreements.

\paragraph{Magnitude of estimated label bias.} To prevent background pixels and large objects from dominating the audit, we use the foreground union of each image, 
%\begin{equation}
$U_i = \max\left\{ 1,\, \left|Y_{\mathrm{obs},i} \cup\widehat Y_i^{\mathrm{CL}}\right| \right\}$, as a normalization term.
%\label{eq:foreground_union} 
%\end{equation}
The image-level Confident Label Error Rates (LER) and group LER are thus:
\begin{equation}
\widehat{\mathrm{LER}}_i = \frac{|\mathcal E_i|}{U_i}, \qquad \widehat{\mathrm{LER}}_g = \frac{1}{|\mathcal I_g|} \sum_{i\in\mathcal I_g} \widehat{\mathrm{LER}}_i, \qquad \mathcal I_g=\{i:g_i=g\}.
\label{eq:sample_ler}
\end{equation}
Averaging image-level rates ensures that each image contributes equally. LER therefore measures foreground-normalized prevalence of potential label bias.
\paragraph{Subgroup concentration.} We quantify the absolute and relative concentration of disagreements as
\begin{equation}
\widehat{\Delta}_{\mathrm{LER}} = \widehat{\mathrm{LER}}_{g_b} - \widehat{\mathrm{LER}}_{g_c},
\qquad
\widehat{\mathrm{RR}} = \frac{ \widehat{\mathrm{LER}}_{g_b}+\varepsilon_{\mathrm{RR}} }{ \widehat{\mathrm{LER}}_{g_c}+\varepsilon_{\mathrm{RR}} },
\label{eq:ler_gap}
\end{equation}
where \(\varepsilon_{\mathrm{RR}}>0\) is a fixed numerical stabilizer. Positive \(\Delta_{\mathrm{LER}}\), equivalently \(\mathrm{RR}>1\), indicates greater CL-identified label bias in \(g_b\).
\paragraph{Direction of label bias.} We partition the potential errors according to whether the observed mask contains foreground excess or deficit relative to CL assignments:
\begin{align}
\mathcal E_i^{\mathrm{exc}} &= \left\{ p\in\mathcal E_i: y_{\mathrm{obs},i,p}=1,\, \widehat y^{\mathrm{CL}}_{i,p}=0 \right\}, \label{eq:excess_set}\\ \mathcal E_i^{\mathrm{def}} &= \left\{ p\in\mathcal E_i: y_{\mathrm{obs},i,p}=0,\, \widehat y^{\mathrm{CL}}_{i,p}=1 \right\}. \label{eq:deficit_set}
\end{align}
Foreground excess is consistent with annotation expansion, whereas foreground deficit is consistent with annotation contraction, relative to the (unknown) true annotation. We define  Directional Rate (DR) as
\begin{equation}
\widehat{\mathrm{DR}}_g = \frac{1}{|\mathcal I_g|} \sum_{i\in\mathcal I_g} \widehat{\mathrm{DR}}_i \quad \text{where} \quad 
\widehat{\mathrm{DR}}_i = \frac{ |\mathcal E_i^{\mathrm{exc}}| - |\mathcal E_i^{\mathrm{def}}| }{U_i},
\label{eq:DR}
\end{equation}
$\widehat{\mathrm{DR}}_g$ lies in \([-1,1]\): positive values indicate foreground excess, negative values indicate foreground deficit, and values near zero indicate either little directional disagreement or cancellation between the two error types. 
% We additionally report $\Delta_{\mathrm{DR}} = \widehat{\mathrm{DR}}^{g_b} - \widehat{\mathrm{DR}}^{g_c}$ to account for baseline directional tendencies of the proxy model. 
For direct interpretation, the corresponding raw group-level counts are 
\begin{equation}
C_{\mathrm{exc}}^g = \sum_{i\in\mathcal I_g} |\mathcal E_i^{\mathrm{exc}}|, \qquad C_{\mathrm{def}}^g = \sum_{i\in\mathcal I_g} |\mathcal E_i^{\mathrm{def}}|.
\label{eq:directional_counts}
\end{equation}
These counts are comparable across methods only when computed over identical evaluation images, resolutions, and folds. To summarize, for label errors, $\widehat{\mathrm{LER}}$ estimates the magnitude, $\widehat{\mathrm{RR}}$ identifies their subgroup concentration, and $\widehat{\mathrm{DR}}$ estimates their direction (under- or over-segmentation). Without independently verified masks, these remain proxy estimates and may also reflect systematic errors or miscalibration in the cross-fitted model. We therefore estimate uncertainty by bootstrapping images, or patients when multiple images belong to the same individual.

\subsection{Mitigating Label Bias via Subgroup Conditioning}
\label{sec:mitigation}

We model group-conditional label bias as an annotation-style shift superimposed on shared anatomical structure~\citep{zepf2023label}. Motivated by the increased subgroup separability observed under controlled corruption in Section~\ref{sec:effects}, our method represents group-specific labeling styles during training and directs all predictions toward a selected reference style at inference. It combines FiLM-based subgroup conditioning with asymmetric masking of potentially biased boundary supervision. We first describe the proposed method and then the comparison baselines.

\subsubsection{Subgroup-Conditioned Decoder (SCD)}
\label{subsubsec:conditioning}

We use FiLM~\citep{perez2018film} to model group-specific annotation styles. Let \(f_i=e_\theta(x_i)\) be the encoder features, with channel \(k\in\{1,\ldots,K_f\}\). An embedding of \(g_i\) produces channel-wise scale and shift parameters \((\gamma_{g_i,k},\delta_{g_i,k})\), which are used to adapt the encoding \(f'_{i,k}=\gamma_{g_i,k}f_{i,k}+\delta_{g_i,k}\), finally yielding \(\widehat y_i=d_\phi(f'_i)\), where \(d_\phi\) is the shared decoder. During inference, every sample is conditioned on the selected reference group: \(\widehat y_i^{\mathrm{ref}} =d_\phi(\gamma_{g_c}\odot e_\theta(x_i)+\delta_{g_c})\). This expresses all predictions in the reference annotation style without requiring the input's group identity at inference.

\paragraph{Asymmetric Boundary Masking and Objective.}
Let \(B_{i,p}=[b_w(y_{\mathrm{obs},i})]_p\in\{0,1\}\) denote a boundary band of width \(w\) extracted by the morphological operator \(b_w\). We assign \(W_{i,p}=1-\mathbf 1[g_i=g_b]B_{i,p}\), removing boundary supervision from \(g_b\) while retaining its interior pixels of the target and all supervision from \(g_c\). FiLM models group-specific annotation variation, while boundary masking prevents the potentially biased group from determining the shared boundary representation. The resulting objective is 
\begin{equation} \mathcal L_{\mathrm{mask}} =
\frac{ \displaystyle\sum_{i=1}^{N}\sum_{p\in\Omega} W_{i,p}\,
\ell_{\mathrm{seg}} (\widehat y_{i,p},y_{\mathrm{obs},i,p}) }{ \displaystyle\sum_{i=1}^{N}\sum_{p\in\Omega}W_{i,p} },
\label{eq:combined}
\end{equation} where \(\ell_{\mathrm{seg}}\) is the pixel-wise segmentation loss. 

\paragraph{Automated reference-group selection (Auto-SCD).}
\label{sec:auto_conditioning}
When group identities are known but the reference group is not, we estimate it from the first \(E_w\) epochs. Motivated by training-dynamics diagnostics and small-loss selection~\citep{swayamdipta2020dataset,li2020dividemix}, we select
\begin{equation}
\widehat g_c = \argmin_{g\in\mathcal G} \frac{1}{E_w|\mathcal I_g|} \sum_{e=1}^{E_w} \sum_{i\in\mathcal I_g} \ell_i^{(e)}.
\label{eq:discovery}
\end{equation}
Training then applies Eq.~\ref{eq:combined} with \(\widehat g_c\) as the reference group, and inference fixes the FiLM condition to \(\widehat g_c\). Because low loss may reflect intrinsic task difficulty rather than annotation quality, we evaluate this heuristic separately from a manually selected reference for SCD. Pseudocode in Appendix~\ref{sec:auto_fim}.

\paragraph{Mitigation Baselines}
% \label{sec:baselines}
 (i) \textbf{Feature alignment} matches group representations using MMD~\citep{gretton2006kernel} or Deep CORAL~\citep{sun2016deep} or adversarial debiasing, which uses a GRL classifier to jointly minimize segmentation loss and suppress group predictability~\citep{ganin2016domain,zhang2018mitigating}. (ii) \textbf{Fairness regularization} penalizes group gaps in mean foreground probability (Demographic Parity, DP)~\citep{dwork2012fairness} or pixel-level true- and false-positive rates computed against the observed masks (Equalized Odds, EO)~\citep{hardt2016equality}. (iii) \textbf{Noise-robust methods} that treat label bias as label noise, including Generalized Cross Entropy (GCE)~\citep{zhang2018generalized} and Bootstrapping~\citep{reed2014training}. Implementation details are given in Appendix~\ref{appendix:mitigation_methods}.

\subsection{Dataset and Experimental Setup}
% \subsection{Dataset \& Experimental Setup}
\label{sec:dataset}

We evaluate our framework on four datasets that progress from controlled label bias to naturally occurring annotation variability. Table~\ref{tab:datasets} summarizes the role of each dataset and the construction of target subgroups. We additionally use the ISIC~\citep{8363547} dataset as exploratory, because of the lack of true annotations for evaluation. 

All models use a U-Net architecture with a pretrained ResNet-34 backbone (ResNet-18 \& SegFormer \texttt{mit-b2} for ablation) on multiple RTX 5000 (16GB). Networks are trained for 15 epochs using the Adam optimizer with an objective of Cross-Entropy and Dice loss. For the unsupervised auto-conditioning phase (Section~\ref{sec:auto_conditioning}), we set $E_w = 5$ epochs to allow stable subgroup loss separation before applying FiLM features. Boundary maps $B(x)$ for the asymmetric spatial loss are extracted on-the-fly via morphological gradient filters with kernel radius $k_r=2$. All experiments use $5$-fold cross-validation to calibrate CL thresholds $\tau_c$. We validate segmentation quality using standard Dice and IoU scores (in \%). Algorithmic fairness is assessed via the subgroup performance gap (we use$\Delta = g_c - g_b$, to denote subgroup differences). The label bias audit uses the CL-derived metrics to assess the true nature of errors (Section~\ref{sec:cl_audit}). For analysis using learned representations, we extract globally pooled features from the encoder bottleneck and quantify how strongly they encode subgroup membership. Full evaluation details are provided in~\ref{app:feature_analysis}.

\begin{table*}[t]
\centering
\small
\setlength{\tabcolsep}{3pt}
\renewcommand{\arraystretch}{1.12}

\begin{tabularx}{\textwidth}{
    >{\hsize=.70\hsize\raggedright\arraybackslash}X
    >{\hsize=.95\hsize\raggedright\arraybackslash}X
    >{\hsize=1.15\hsize\raggedright\arraybackslash}X
    >{\hsize=1.45\hsize\raggedright\arraybackslash}X
    >{\hsize=.75\hsize\raggedright\arraybackslash}X
}
\toprule
\textbf{Dataset} &
\textbf{Data} &
\textbf{Label variation} &
\textbf{Group construction} &
\textbf{Setting} \\
\midrule

\textbf{CelebAMask-HQ (CMask)}
\par{\scriptsize\citep{CelebAMask-HQ}}
&
30,000 face images with semantic masks and a binary male/female attribute
&
Boundary-localized erosion at
\(\beta\in\{0,25,50,75,100\}\%\), following
Section~\ref{sec:formalizing}
&
\(g_c\): males with unmodified masks.
\(g_b\): females, with a proportion \(\beta\) receiving eroded masks
(\(r_d=15\), unless otherwise specified)
&
Controlled: error location, prevalence, and direction are known
\\

\addlinespace[2pt]

\textbf{PhC-U373}
\par{\scriptsize\citep{mavska2014benchmark,ulman2017objective}}
&
651 glioblastoma-cell images with two annotation styles
&
Style~0 provides detailed, tight boundaries; style~1 provides coarser,
expanded boundaries
&
\(g_c\): untinted images retaining style~0.
\(g_b\): synthetically tinted images, with a proportion \(\beta\) receiving
style~1 annotations
&
Semi-synthetic: natural label styles with subgroup assignment
\\

\addlinespace[2pt]

\textbf{IMA++}
\par{\scriptsize\citep{abhishek2025imaplusplus}}
&
2,394 dermoscopic images (ISIC derived) with two to five annotations per image; skin tone is
estimated as ST1--ST4
&
Naturally occurring detailed/tight and loose/expanded annotation styles across
annotators
&
\(g_c\): lighter-tone images (ST1-2).
\(g_b\): darker-tone images (ST3-4).
At \(\beta=0\), natural bias is retained.
At \(\beta\in\{50,100\}\%\), \(\beta\) of \(g_b\) is reassigned annotations from loose annotators, while \(g_c\) unchanged
&
Natural annotations with a semi-synthetic group--style association
\\

\addlinespace[2pt]

\textbf{ISIC}
\par{\scriptsize\citep{8363547}}
&
Dermoscopic lesion images with one observed segmentation mask per image; skin
tone is estimated as ST1--ST4
&
Naturally occurring subgroup-associated differences in observed segmentation
masks; no alternative clean or reference annotation is available
&
\(g_c\): lighter-tone images (ST1--ST2).
\(g_b\): darker-tone images (ST3--ST4).
No annotation-style reassignment or synthetic corruption is applied
&
\textbf{Exploratory}: observed disparities cannot be uniquely attributed to label error
\\

\bottomrule
\end{tabularx}

\caption{
\textbf{Datasets and group construction to audit label bias.} Further detailed in Appendix~\ref{app:dataset_detailed}.
}
\label{tab:datasets}
\end{table*}

% \subsection{Experimental Setup}

\section{Results and Discussion}

\subsection{Effect of Label Bias on Learned Representations}
\label{sec:effects}

\begin{figure}[t]
    \centering
    \includegraphics[width=\linewidth]{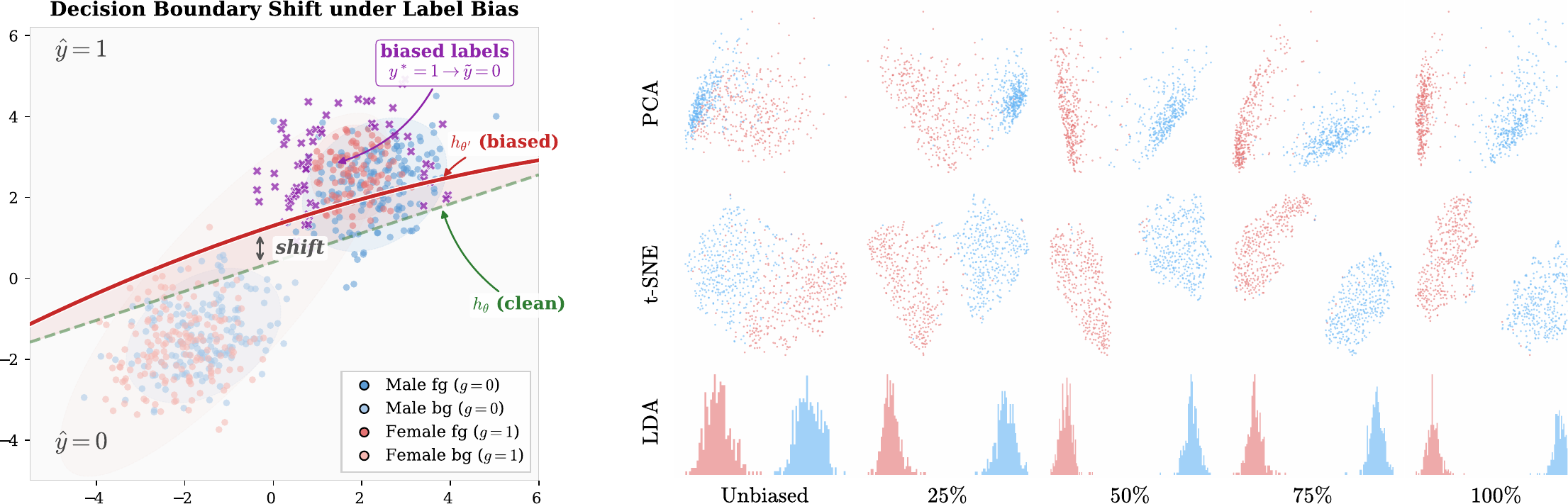}
  \caption{(Left) Biased labels (purple) shift the learned decision boundary  $h_{\theta'}$ away from the clean boundary $h_\theta$, producing systematically incorrect segmentation during inference. (Right) Effects of increasing label bias on subgroup separability in the encoder latent space (CelebAMask-HQ). Red: $g_b$; blue: $g_c$. Rows show PCA (top), t-SNE (middle), and LDA (bottom).
    }
    \label{fig:effects}
    % \vspace{-10pt}
\end{figure}

When membership of the group $g_b$ is encoded in the data, label bias degrades boundary delineation for the affected subgroup ($g_b$) while leaving $g_{c}$ performance largely unaffected~\citep{vorontsov2021label,li6017137noise}, as illustrated in Fig~\ref{fig:effects}~(Left). \citet{jones2023role} demonstrated that if groups are not well separable in feature space, the model cannot limit the bias to just $g_b$, and errors may propagate across both groups, creating a large fairness gap. This shows that \textbf{subgroup separability governs whether label bias appears as a subgroup disparity or an overall performance degradation}, which may be indistinguishable through group-level metrics.

Through controlled experiments, we show that as \(\beta\) increases from \(0\%\) to \(100\%\), the bottleneck Maximum Mean Discrepancy (MMD) rises from \(0.10\) to \(0.29\) and centroid distance from \(7.54\) to \(19.04\) (Fig.~\ref{fig:effects}~(Right)), alongside the PCA, t-SNE, and LDA visualizations. The separation at $\beta=0$ might reflect genuine visual differences, whereas growth under increasing systematic corruption indicates that label bias amplifies subgroup separation (see further analysis in Appendix~\ref{app:feature_analysis}). This extends prior classification evidence that systematic label bias encourages models to encode sensitive attributes along with subgroup separability~\citep{stanley2025exploring,jones2023role}.

Subgroup separability by itself is \textit{neither a fairness violation nor an undesired property}~\citep{lahoti2020fairness,petersen2023demographically}. In domains such as medical imaging, where disease manifestation may differ across populations and relevant structures may genuinely differ across subgroups, enforcing marginal representational invariance can remove useful information or promote equal risk, also known as the independence fairness criterion~\citep{petersen2023demographically, fairmlbook}. Our results identify label bias as an underappreciated confounder of such separability, and suppressing the representations via feature alignment and fairness objectives to treat label bias does not outperform the ERM baseline and can cause \textit{leveling down}~\citep{zong2023medfairbenchmarkingfairnessmedical, zietlow2022leveling}. We therefore treat increased separability as a supervision signal and model this variation before mapping all outputs to a selected reference annotation style in Section~\ref{subsubsec:conditioning}.

% \subsection{Detecting Label Bias via Confident Learning}

\subsection{Standard Metrics Fail to Detect Label Bias}
\label{sec:metrics_failure}

\begin{figure}[t]
    \centering
    \includegraphics[width=\linewidth]{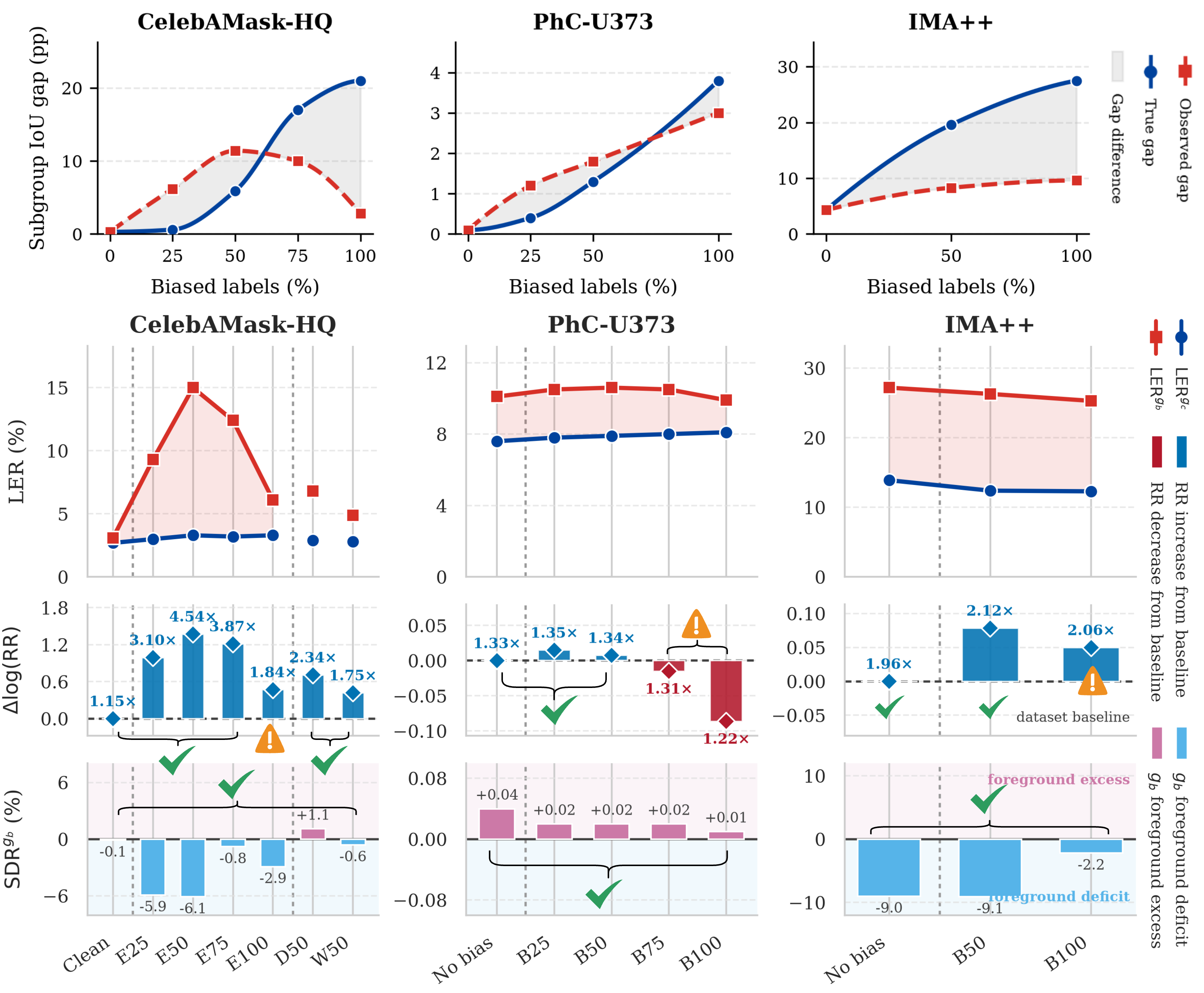}
    \caption{
    \textbf{Label-bias evaluation. NOTE: True performance against clean labels is unavailable in practice and provided here solely for validation.} 
    Top: IoU gaps computed against clean and observed masks. Bottom: CL audit using observed masks only; $\widehat{LER}$ measures suspected error prevalence, $\widehat{RR}$ its relative concentration in $g_b$, and $\widehat{DR}$ its direction. }
    \label{fig:unified_label_bias_audit}
    \vspace{-6pt}
\end{figure}

As shown in Fig.~\ref{fig:unified_label_bias_audit}~(Top), across all datasets and conditions, \textbf{IoU consistently mistakes the magnitude of true subgroup disparity when evaluated against observed labels}. At peak, on CelebA $\beta=50\%$, the IoU gap for observed performance is 11.4\%, which reduces to 10.0\% at $\beta=75\%$, suggesting improved fairness, while the true gap shows severe and increasing disparity (5.5\% $\rightarrow 17.0\%$). This demonstrates the \textit{biased ruler effect}: metrics evaluated against corrupted labels not only fail to detect bias but also actively hide it~\citep{parikh2025investigating}. A similar trend is also observed on PhC-U373 (a 2.4\% gap between true and observed performance at $\beta=100\%$) and IMA++, where observed IoU remains nearly constant even as label corruption increases. 

The observed gap \textit{overestimates} the true gap for dilation (at $\beta=50\%$~$\Delta_{\text{obs}}{=}8.28$ vs.\ $\Delta_{\text{true}}{=}1.65$), while underestimating it for erosion at the same level. This \textbf{asymmetry is a fundamental property of Dice}: erosion removes true positives, which appear in both the numerator and denominator $\frac{2\text{TP}}{2\text{TP} + \text{FP} + \text{FN}}$, whereas dilation adds false positives that inflate only the denominator. Thus, the same number of corrupted pixels produces a larger drop in Dice under erosion than dilation, making observed metrics unreliable for comparing bias severity.

\begin{quote}
    \textit{This calls for better evaluation metrics, as any fairness intervention based on closing this observed gap could therefore increase the bias, not reduce it.}
\end{quote}

\subsection{Auditing Group-Conditional Label Bias Without Ground-Truth}
\label{sec:audit_results}

Fig.~\ref{fig:unified_label_bias_audit} summarizes the audit and CL-metrics across all datasets under various conditions. Detailed results for the audit are in Appendix~\ref{appendix:detailed_mitigation}. \textbf{{While our primary contribution is auditing without ground-truth, a small set of clean labels for validation provides direct confirmation of detected bias; refer to Appendix~\ref{sec:cl_clean} and Table~\ref{tab:trainclean_results}.}}

\textbf{On CelebAMask-HQ,} \(\mathrm{LER}^{g_c}\) remains stable between \(2.7\%\) and \(3.3\%\), whereas \(\mathrm{LER}^{g_b}\) increases from \(3.1\%\) without corruption to \(9.3\%\) and \(15.0\%\) at \(25\%\) and \(50\%\) erosion. Consequently, $\text{log}\mathrm{(RR)}$ stay $>0$, correctly identifying that suspected errors are concentrated in \(g_b\). Erosion also produces negative DR, reaching \(-6.1\%\) at \(50\%\) corruption, while dilation produces positive DR (\(+1.1\%\)). The audit therefore detects both the subgroup concentration and expected direction of label errors. However, at complete erosion, RR decreases, and DR weakens to \(-2.9\%\). This attenuation is a fundamental limitation of CL-based auditing without ground-truth, which remains effective for moderate bias, the most common real-world case, where bias is systematic but not whole.

\textbf{PhC-U373} presents a very subtle case of labeling bias (affecting just $\approx$4\% of pixels) due to visual domain shift. The LER of both groups remains nearly constant, while \(\mathrm{LER}^{g_b}\) changes from \(10.1\%\) to \(9.9\%\). DR remains positive, indicating over-segmentation of $g_b$. Thus, the audit identifies a persistent baseline difference between the subgroups. \textbf{IMA++} exhibits a large disparity before additional corruption: baseline LER is \(13.9\%\) for \(g_c\) and \(27.2\%\) for \(g_b\), corresponding to \(\mathrm{RR}=1.96\), and DR $\leq$ 0, indicating under-segmentation of $g_b$. The baseline-adjusted RR signal, along with baseline LER \& DR,  indicates that pre-existing differences in image difficulty or annotation style dominate the additional corruption signal. 
\textbf{On ISIC,} with no artificial bias, the audit detected a naturally elevated error rate, confirming our evaluation on a real-world clinical dataset with no clean validation.

% \begin{table*}[h]
% \centering
% \resizebox{\textwidth}{!}{
% \begin{tabular}{l|ccc|ccc|ccc|ccc}
% \toprule
%  & \multicolumn{3}{c|}{\textbf{True IoU}} & \multicolumn{3}{c|}{\textbf{Observed IoU}} & \multicolumn{3}{c|}{\textbf{LER (CL Audit)}} & \multicolumn{3}{c}{\textbf{DR (CL Audit)}} \\
% \textbf{Experiment} & \textbf{Clean} & \textbf{Biased} & \textbf{Gap} & \textbf{Clean} & \textbf{Biased} & \textbf{Gap} & \textbf{Clean} & \textbf{Biased} & \textbf{Gap} & \textbf{Clean} & \textbf{Biased} & \textbf{Gap} \\
% \midrule
% No Bias ($0\%$) & 0.892 & 0.849 & 0.043 & 0.892 & 0.849 & 0.043 & 0.139 & 0.272 & 0.134 & -0.066 & -0.090 & 0.023 \\
% Bias $50\%$ & 0.902 & 0.706 & 0.196 & 0.902 & 0.819 & 0.083 & 0.124 & 0.263 & 0.139 & -0.042 & -0.091 & 0.049 \\
% Bias $100\%$ (Baseline) & 0.902 & 0.627 & 0.275 & 0.902 & 0.806 & 0.096 & 0.123 & 0.253 & 0.130 & -0.034 & -0.022 & -0.012 \\
% \bottomrule
% \end{tabular}
% }
% \caption{IMA++ Label Bias Results. Evaluating True/Observed IoU alongside Confident Learning's Label Error Rate (LER) and Signed Directional Rate (DR). A negative DR indicates that the training annotations tend to be smaller/eroded compared to the confident model predictions.}
% \label{tab:imapp_results_full_DR}
% \end{table*}

CL metrics such as LER and RR can identify label bias when the corrupted pattern remains distinguishable from the underlying image signal. At the same time, DR can recover whether annotations exhibit foreground deficit or excess. Their sensitivity decreases when the proxy model learns systematic errors or when pre-existing subgroup differences dominate prediction-annotation disagreement. The metrics should therefore be interpreted as indicators of potential label bias rather than direct estimates of true annotation error.

\subsection{Mitigating Bias via Subgroup Conditioning (SCD)}

Under label corruption, fairness constraints, alignment methods, and standard label noise methods fail to close the fairness gap, as shown in the results in Table~\ref{tab:mitigation_results}. The proposed SCD methods close the IoU gap to 8.9\% from 21.0\% at baseline, for $\beta = 100\%$. Ablations in Appendix~\ref{app:labelnoise_baselines}. 

These results help distinguish label bias from conventional fairness problems and label noise. Systematic label corruption is much more complex because it corrupts the (learned) input representation, while traditional fairness approaches ($\mathrm{EO/DP}$ regularization) constrain the model outputs, which do not target the source of subgroup boundary shifts. Likewise, alignment methods try to suppress (potentially crucial) subgroup information via invariance. Our approach, instead, retains subgroup information and applies a cleaner reference condition at inference. 

Even though subgroup discovery identified clean reference labels in all conditions (see Appendix~\ref{app:autoscd} for robustness test for Auto-SCD), \textbf{we recommend} auditing for label bias (Section~\ref{sec:cl_audit}) and then referencing the clean(er) subgroup after inspection or expert validation. Auto-conditioning is particularly useful when subgroups are unknown.

\begin{table}[t]
\centering
\caption{
\textbf{Comparison of label-bias mitigation methods.}
For each setting, $(\text{IoU}/\Delta)_{\text{true}}$ reports mean IoU / subgroup IoU gap ($\Delta$) against true-reference masks (in \% and percentage points). CL-metrics report foreground excess/deficit counts (E/D in millions of pixels). Rows with \textbf{Ours} are in gray.
}
\label{tab:mitigation_results}
\scriptsize
\setlength{\tabcolsep}{1.6pt}
\renewcommand{\arraystretch}{0.90}

\resizebox{\linewidth}{!}{%
\begin{tabular}{@{}ll cc cc c c@{}}
\toprule
& & \multicolumn{2}{c}{\textbf{CelebA ($\beta=50\%$)}} & \multicolumn{2}{c}{\textbf{CelebA ($\beta=100\%$)}} & \textbf{PhC-U373} & \textbf{IMA++ ($\beta=50\%$)} \\
\cmidrule(lr){3-4} \cmidrule(lr){5-6} \cmidrule(lr){7-7} \cmidrule(lr){8-8}
\textbf{Category} & \textbf{Method} 
& \textbf{$(\text{IoU/$\Delta$})_\text{true}$} & \textbf{CL (E/D)} 
& \textbf{$(\text{IoU/$\Delta$})_\text{true}$} & \textbf{CL (E/D)} 
& \textbf{$(\text{IoU/$\Delta$})_\text{true}$}
& \textbf{$(\text{IoU/$\Delta$})_\text{true}$} \\
\midrule
\textit{Reference} & ERM 
& $93.1/5.5$ & $39.3/86.4$ & $82.5/21.0$ & $9.0/26.9$ & $91.2/4.4$ & $81.7/19.6$ \\
\midrule
\multirow{3}{*}{\textit{Fairness}} 
& EO & $93.0/5.7$ & $33.1/79.7$ & $82.6/21.0$ & $9.0/26.9$ & $91.9/3.6$ & $79.8/13.9$ \\
& DP & $93.0/5.6$ & $34.0/78.5$ & $82.6/21.0$ & $8.9/26.9$ & $92.0/3.4$ & $79.6/13.6$ \\
& EO+DP & $92.8/5.8$ & $35.1/77.1$ & $82.4/20.9$ & $10.4/26.7$ & $91.9/3.5$ & $79.9/13.1$ \\
\midrule
\multirow{3}{*}{\textit{Alignment}} 
& Adversarial & $92.4/6.7$ & $37.0/76.9$ & $82.7/21.1$ & $3.6/10.7$ & $91.8/3.7$ & $80.3/13.9$ \\
& MMD & $92.8/6.3$ & $34.2/81.2$ & $82.8/20.5$ & $3.7/12.3$ & $91.7/3.9$ & $80.0/13.0$ \\
& CORAL & $93.2/5.5$ & $32.3/80.6$ & $82.6/20.9$ & $8.7/26.9$ & $91.7/3.9$ & $79.7/13.9$ \\
\midrule
\multirow{3}{*}{\textit{Noise}} 
& GCE & $92.3/6.6$ & $36.9/77.6$ & $82.5/21.3$ & $9.2/26.0$ & $91.0/3.9$ & $80.0/14.6$ \\
& Bootstrapping & $92.8/5.8$ & $42.0/83.9$ & $82.5/21.1$ & $8.9/26.8$ & $91.0/3.7$ & $80.2/14.7$ \\
& Asym. mask & $91.5/7.9$ & $39.6/61.0$ & $79.8/24.7$ & $3.6/31.0$ & $90.9/5.0$ & $79.7/14.6$ \\
\midrule
\rowcolor{gray!15}
& SCD & $96.8/0.5$ & $10.4/95.4$ & $89.0/8.9$ & $2.5/92.7$ & $91.8/3.6$ & $85.8/6.7$ \\
\rowcolor{gray!15}
& SCD+Asym & $\mathbf{96.8/0.5}$ & $2.5/23.7$ & $\mathbf{89.2/9.0}$ & $2.6/93.3$ & $91.1/4.8$ & $86.0/7.0$ \\
\rowcolor{gray!15}
\multirow{-3}{*}{\textbf{Ours}} 
& Auto-SCD & $93.7/4.2$ & $28.2/86.8$ & $85.3/16.1$ & $4.6/45.2$ & $\mathbf{92.5/3.2}$ & $\mathbf{87.4/6.2}$ \\
\bottomrule
\end{tabular}%
}
\vspace{-6pt}
\end{table}

\section{Limitation and Intended Scope}

{As for our limitations,} the CL-based audit estimates prediction-annotation disagreement as a signal for label bias. Without independently verified annotations, disagreement may also reflect model error, miscalibration, or genuine differences in task difficulty. These disagreements underestimate severe bias when models learn the corrupted annotations (like in our $\beta$ = 100\% condition). The mitigation method assumes that a subgroup with a different annotation style can be identified. The training loss heuristic in auto-discovery (Section~\ref{sec:auto_conditioning}) may reflect differences in imaging difficulty or anatomy. If all groups are biased, the conditioning may reproduce, as expected, undesirable biased predictions (Appendix~\ref{appendix:conditioning_ablations}). Reference selection should therefore be supported by expert assessment or a verified subset. We test and report counterfactual and calibration ablations, the no-biased-subgroup case, and model-capacity dependence in Appendix~\ref{appendix:conditioning_ablations}. 
Finally, label bias simulation (Section~\ref{sec:formalizing}) captures only a subset of possible annotation error types, despite our efforts to make them realistic and complex. Most segmentation datasets lack verified clean masks, like our exploratory dataset, ISIC; therefore, we rely on controlled simulation and use verified masks for validation, where possible. 

\section{Conclusion}

% \textbf{To conclude},
We introduce a framework for simulating and auditing label bias for segmentation without ground-truth labels. Our proposed subgroup-conditioned (SCD) mitigation strategy reduces subgroup disparities, where traditional debiasing and denoising methods fail. We also address a blind spot in algorithmic fairness in image segmentation, showing that biased labels can both distort fairness evaluation and learned representations. Our proposed CL-metrics detect subgroup concentration and direction of these errors, where standard segmentation metrics fail, possibly presenting a misleading narrative. 

\textbf{As future work}, we focus on directly addressing current limitations by extending this framework to measure under severe bias situations, relaxing the need to know subgroup membership for mitigation and validation on datasets with more naturally occurring label bias.

\clearpage

\subsection*{Generative AI Disclosure}

Following fair and responsible use of AI for research, we list the use of generative AI for the following sub-tasks:

\begin{itemize}
    \item \textbf{Major use:} Formatting and designing \LaTeX{} tables for results (manually audited), writing boiler-plate code for pipelines (verified against official implementations), restructuring appendix sections for readability, and compiling anonymous-code repository.
    \item \textbf{Minor use:} Inspiration for figure/plot designs, polishing select paragraphs in the appendix for clarity, and experiment logging.
    \item \textbf{Not used for:} Ideation, developing the methodology, writing the main text (introduction, methods, results, and discussion), mathematical proofs, paper's structure, or in bibliography.
\end{itemize}

All AI-assisted outputs were manually verified by the authors. The authors take full responsibility for the final output, and claims in this work.



\bibliography{iclr2027_conference}
\bibliographystyle{plainnat}

\clearpage

\appendix
\section{Bias Construction and Morphological Operations}

\subsection{Additional Details: Morphological Operations \& Effect of $r_{d}$}

\begin{figure}[!htbp]
    \centering

    \begin{subfigure}{0.8\textwidth}
        \centering
        \includegraphics[width=\textwidth]{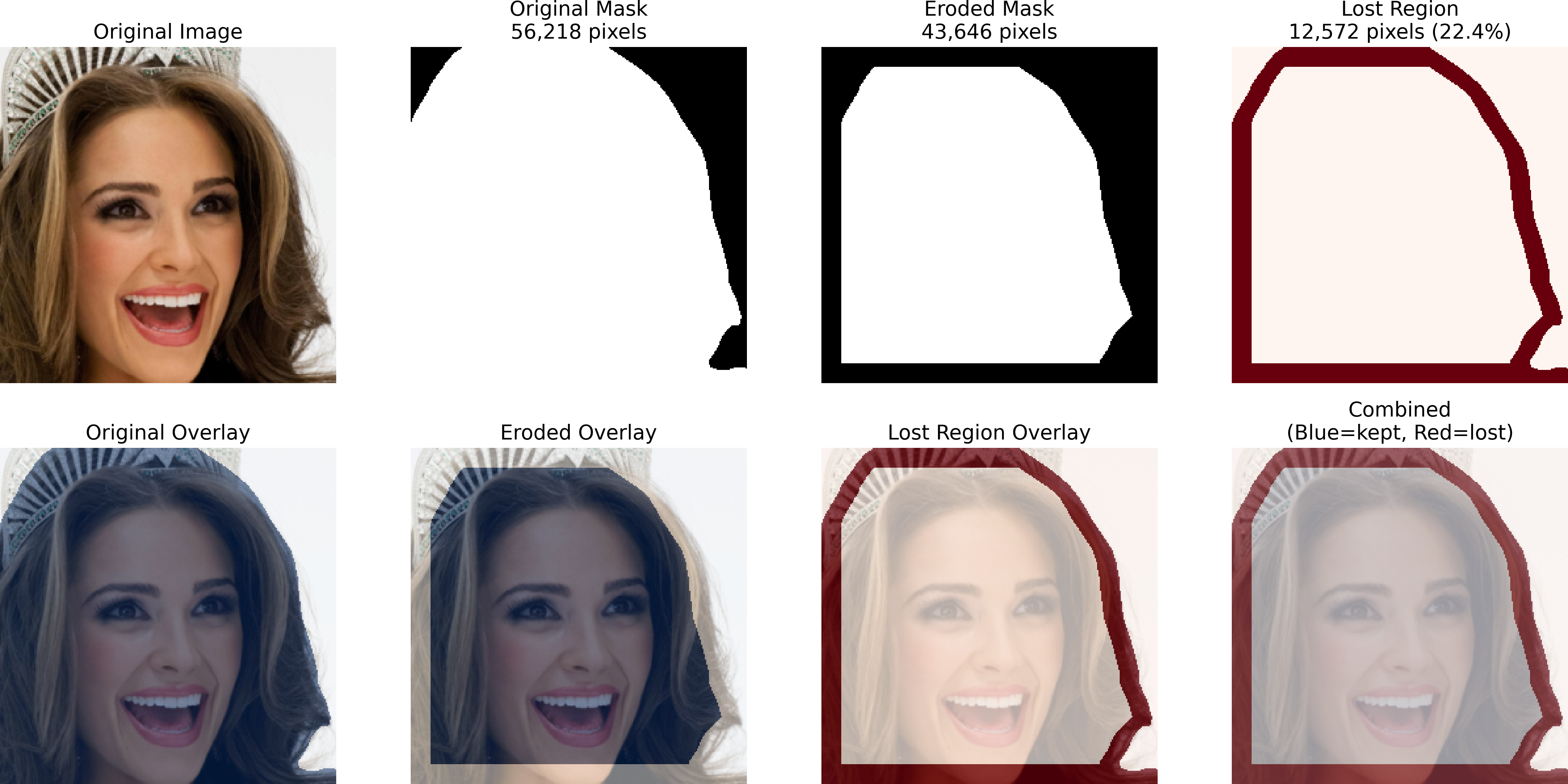}
        \caption{\textbf{A visualization of the erosion operation.} The original ground-truth mask is reduced, isolating the 'Lost Region' (highlighted in red). The combined overlay demonstrates how the erosion strictly affects the boundary, mapping precisely to under-segmentation.}
        \label{fig:erosion_top}
    \end{subfigure}

    \vspace{0.5em}

    \begin{subfigure}{\textwidth}
        \centering
        \includegraphics[width=\textwidth]{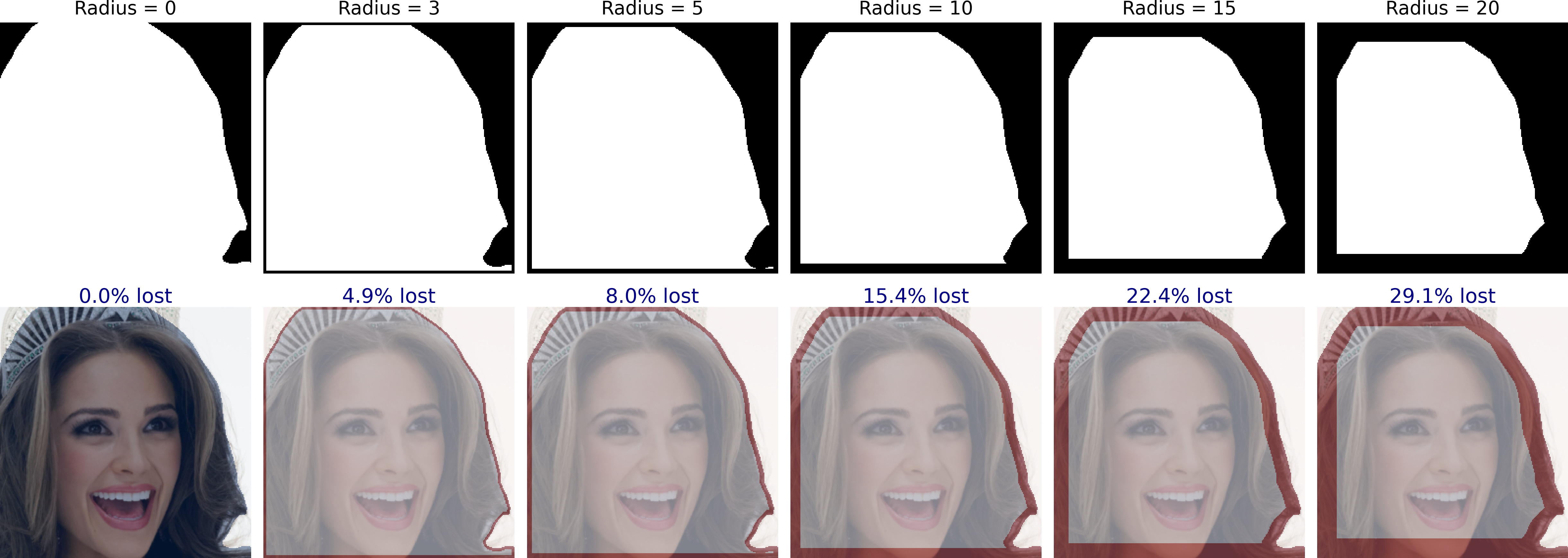}
        \caption{ The visual impact of scaling the erosion kernel radius ($r_d$) from $0$ (Unbiased) to $20$ pixels. The corresponding percentage of foreground pixel loss is tracked for the individual sample, highlighting the severity of boundary retraction at higher values.}
        \label{fig:erosion_bottom}
    \end{subfigure}

    \begin{subfigure}{\textwidth}
        \centering
        \includegraphics[width=\textwidth]{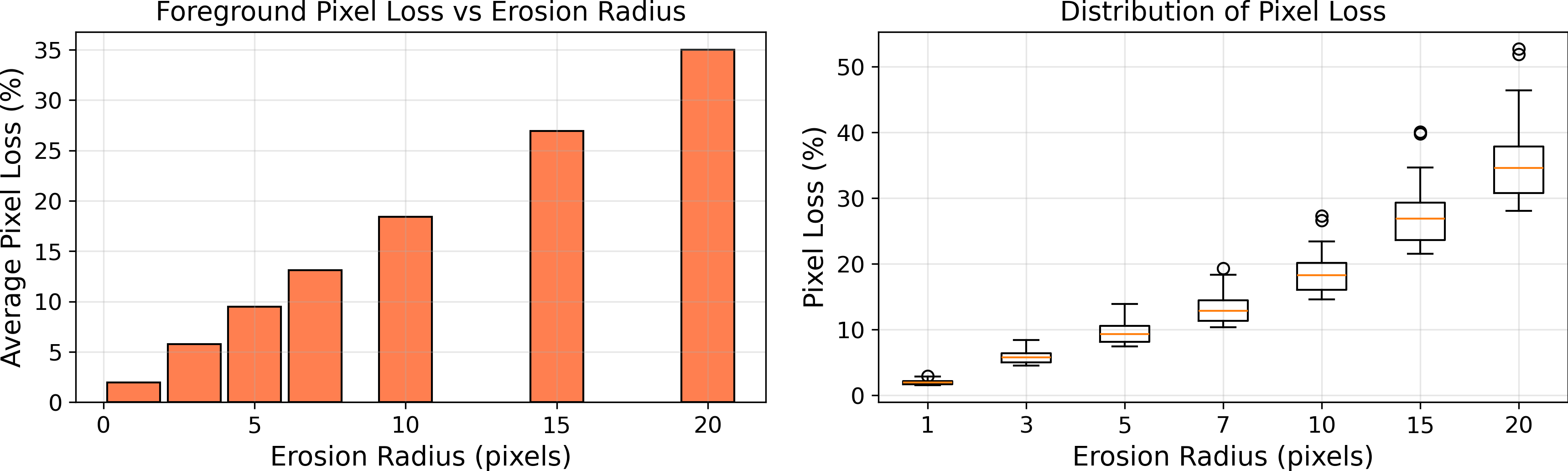}
        \caption{\textbf{Quantitative evaluation:} The bar chart on the left displays the average \% of foreground pixel loss as a function of the $r_e$ (erosion radius). The box plot on the right illustrates the distribution and variance of pixel loss.}
        \label{fig:erosion_stats}
    \end{subfigure}

    \caption{Visual and statistical analysis of synthetic bias (morphological erosion) applied to the CelebAMask-HQ dataset.}
    \label{fig:erosion_comparison}
\end{figure}

Refer to Figure~\ref{fig:erosion_comparison}~(a) for visualization of the representative erosion operation.

\paragraph{Harmonic Boundary Deformation (HBD).}
\label{appendix:hbd}

a superposition of $H{=}3$ sinusaoidal harmonics with randomized
frequencies $\omega_h$, orientations $\alpha_h$, and phases
$\psi_h$:
\begin{equation}
    d(p) = \frac{\rho}{H} \sum_{h=1}^{H}
    \sin\!\bigl(\omega_h\,(x_p \cos\alpha_h + y_p \sin\alpha_h)
    + \psi_h\bigr),
\end{equation}

where $\rho$ controls the maximum displacement amplitude. The altered mask is obtained by thresholding: $\tilde{y}_p = \mathbbm{1}[\phi(p) > d(p)]$. This produces boundaries that locally oscillate between erosion and dilation, simulating the non-uniform annotation inconsistencies observed in practice.

% \clearpage

\section{Representation Analysis and Datasets}

\subsection{Additional Details: Effect of Label Bias on Group Separability}
\label{app:feature_analysis}

Here, we investigate the effect of label bias on the learned representations (discussed in Section~\ref{sec:effects}) in detail. The following is an empirical analysis for feature separability:
\paragraph{Feature Extraction.}
For each saved segmentation model checkpoint, we extracted representations from the final layer of the encoder (e.g, ResNet-50). To obtain a finite-length embedding vector for each image, we applied Global Average Pooling (GAP) to the spatial dimensions of the feature map, resulting in a 1-D bottleneck feature vector.
\paragraph{Visualizations.}
To visualize the high-dimensional feature embeddings, we use three standard projection techniques:
\begin{itemize}[leftmargin=1.0em]
    \item \textbf{t-SNE~\citep{van2008visualizing}:} Computed with 2 components, perplexity set to 30, and initialized via PCA (Principal Component Analysis).
    \item \textbf{PCA~\citep{pearson1901liii}:} Standard Principal Component Analysis projected to the first 2 principal components.
    \item \textbf{LDA~\citep{fisher1936use}:} Linear Discriminant Analysis projected to 1 component, showcasing the maximum linear separability between the subgroup distributions in feature space.
\end{itemize}
\paragraph{Separability Metrics}
To  quantify the group separability effect in the feature space, we computed the following metrics:
\begin{itemize}[leftmargin=1.0em]
    \item \textbf{Linear Probe~\citep{alain2018understanding}~(Accuracy \& AUROC):} We train a logistic regression classifier (LBFGS solver, max iterations = $1000$) on the frozen GAP features to predict subgroup membership. We report the mean accuracy and Area Under the ROC Curve (AUROC) via 5-fold cross-validation.
    \item \textbf{Silhouette Score~\citep{rousseeuw1987silhouettes}:} Evaluates the cluster cohesion and separation based on subgroup labels using Euclidean distance.
    \item \textbf{Fisher's Discriminant Ratio~\citep{mika1999fisher}:} Measures the ratio of between-group variance to within-group variance across the feature dimensions. Statistical significance was verified using a permutation test ($500$ to $1000$ permutations).
    \item \textbf{Maximum Mean Discrepancy (MMD)~\citep{mika1999fisher}:} Computes the distributional distance between the subgroups' features using a multi-scale Gaussian (RBF) kernel.
    \item \textbf{Hotelling's $T^2$ Test~\citep{hotelling1992generalization}:} A multivariate statistical test for differences in the mean feature vectors (centroids) between the two subgroups. If the feature dimensionality exceeded sample size constraints, features were first reduced via PCA to a maximum of 50 dimensions before applying the test.
\end{itemize}

Refer to Table~\ref{tab:feature_separability} for detailed separability analysis with varying $\beta$ on CelebAMask-HQ and Table~\ref{tab:phc_feature_separability} on PhC-U373 and IMA++ dataset. We refrain from feature analysis on ISIC due to lack of clean verified labels.

The linear probe and five unsupervised metrics quantify how well the encoder's feature clusters subgroup membership. For \textbf{analysis on CelebAMask-HQ}, at the baseline ($\beta{=}0$), moderate separability exists due to genuine visual correlates of gender. As we induce label bias by increasing $\beta$, every metric increases: Silhouette score increases $3.9\times$ ($0.137 \to 0.536$), Fisher discriminant ratio $8.8\times$
($0.35 \to 3.10$), and centroid distance $2.5\times$
($7.54 \to 19.04$) from baseline to $\beta{=}100\%$. This confirms the visual claims in Fig.~\ref{fig:effects} (Right) that label bias amplifies subgroup separability beyond just visual differences. All $T^2$ tests reject the null of equal group centroids ($p < 0.001$). \textbf{Results are consistent on PhC-U373 \& IMA++ dataset}, where separability increases with $\beta$, with probe accuracy remaining constant at $1.0$, also at baseline. This is expected, as we add a light synthetic color tint, which introduces a low-level domain shift that dominates the feature space. This makes these two groups perfectly separable even at baseline (i.e, at no added label bias), irrespective of annotation quality. The annotation-style bias induced due to annotator shift constitutes a small fraction of the representation (4\% pixel shift), and its effects are masked by already well-saturated visual separation due to the synthetic tint added. \textbf{On IMA++}, subgroup separability increases with the induced label bias. From $\beta=0\%$ to $100\%$, the silhouette score increases from 0.041 to 0.098 and the centroid distance from 6.366 to 9.629. The Fisher ratio is non-monotonic at $\beta=50\%$, but reaches its highest value of 1.074 at $\beta=100\%$. These results suggest that annotation style, along with skin-tone group, strengthens the encoding of group-related information, although the low absolute silhouette scores indicate overlap between the subgroup representations.

\begin{table*}[t]
\centering
\caption{\textbf{Effect of label bias on representation separability (CelebAMask-HQ).}
Separability increases with bias $\beta$, indicating that label bias amplifies subgroup separability beyond visual differences.}
\label{tab:feature_separability}
\setlength{\tabcolsep}{4pt}
\small

\begin{tabular}{l cc cc cc c}
\toprule
\multirow{2}{*}{\textbf{$\beta$}}
& \multicolumn{2}{c}{\textbf{Supervised}}
& \multicolumn{2}{c}{\textbf{Cluster}}
& \multicolumn{2}{c}{\textbf{Distance}}
& \textbf{Sig.} \\
\cmidrule(lr){2-3} \cmidrule(lr){4-5} \cmidrule(lr){6-7}
& Acc. & AUC
& Sil. & Fisher
& MMD & $\Delta_c$
& $p$ \\
\midrule
0\%   & 0.929 & 0.979 & 0.137 & 0.352 & 0.095 & 7.54  & $<10^{-16}$ \\
25\%  & 0.968 & 0.996 & 0.346 & 1.292 & 0.178 & 13.42 & $<10^{-16}$ \\
50\%  & \textbf{0.980} & \textbf{0.998} & 0.412 & 1.818 & 0.153 & 16.84 & $<10^{-16}$ \\
100\% & 0.979 & 0.997 & \textbf{0.536} & \textbf{3.100} & \textbf{0.292} & \textbf{19.04} & $<10^{-16}$ \\
\midrule
\multicolumn{1}{r}{\footnotesize $\times$ increase}
& \footnotesize 1.05 & \footnotesize 1.02
& \footnotesize 3.91 & \footnotesize 8.81
& \footnotesize 3.07 & \footnotesize 2.52
& \\
\bottomrule
\end{tabular}
\end{table*}

\begin{table}[b]
\centering
\caption{
\textbf{Feature Separability Analysis across Bias Levels.}
Evaluation of the encoder's ability to separate demographic subgroups in the latent feature space for PhC-U373 and IMA++ datasets. Higher values indicate stronger subgroup separability in the learned feature space.
}
\label{tab:phc_feature_separability}
\small
\begin{tabular}{@{}lccc@{}}
\toprule
\textbf{Bias Level ($\beta$)} & \textbf{Silhouette} & \textbf{Fisher Ratio} & \textbf{Centroid Dist.} \\
\midrule
\multicolumn{4}{c}{\textbf{PhC-U373}} \\
\midrule
0\% & 0.217 & 0.614 & 19.631 \\
25\% & 0.209 & 0.555 & 19.102 \\
50\% & 0.208 & 0.517 & 19.648 \\
75\% & 0.192 & 0.495 & 18.006 \\
100\% & 0.259 & 0.718 & 21.549 \\
\midrule
\multicolumn{4}{c}{\textbf{IMA++}} \\
\midrule
0\% & 0.041 & 0.868 & 6.366 \\
50\% & 0.086 & 0.798 & 8.443 \\
100\% & 0.098 & 1.074 & 9.629 \\
\bottomrule
\end{tabular}
\end{table}

\subsection{Additional Details: Extended Dataset Descriptions}
\label{app:dataset_detailed} 

\subsubsection{PhC-U373}
\label{appendix:phc}
This dataset was part of the ISBI Cell tracking challenge~\citep{ulrich2009mechanical,ulman2017objective}, which here gives us a unique opportunity to evaluate bias mitigation in the presence of multi-expert annotations. The specific dataset is curated by~\citet{zepf2023label} using the first video sequence, which contains 115 2D images of multiple cells annotated with two classes: Cells and Background. It contains 651 images of single cells resized to 128x128 pixels. Annotators were instructed to perform either detailed annotations or coarse/wider annotations (on average 4\% bigger by volume).

\begin{figure}[!htbp]
    \centering
    \begin{subfigure}[b]{0.41\textwidth}
        \centering
        \includegraphics[width=\textwidth]{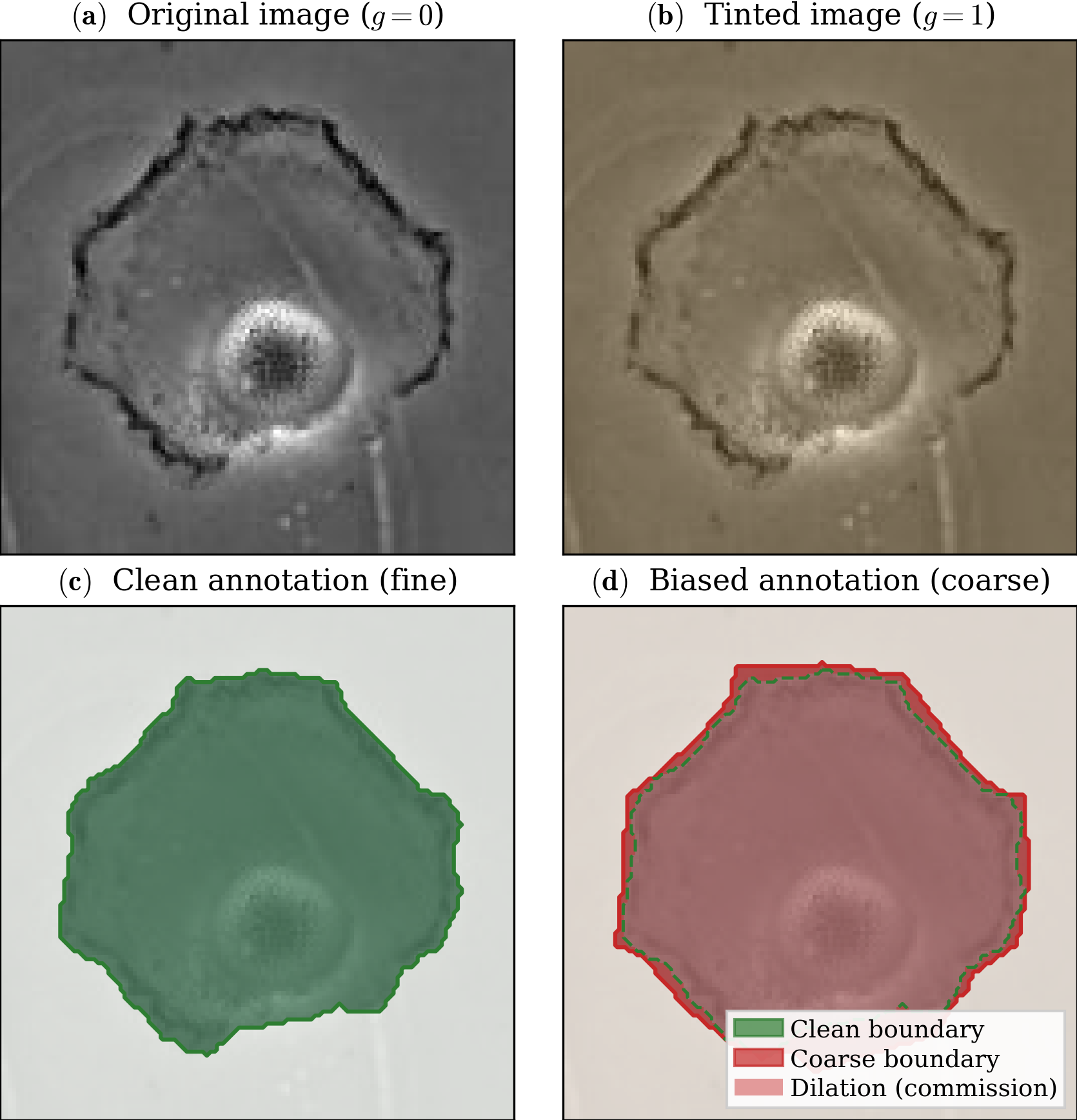}
        % \caption{Low bias setting}
        % \label{fig:effects_a}
    \end{subfigure}
    \hfill
    \begin{subfigure}[b]{0.57\textwidth}
        \centering
        \includegraphics[width=\textwidth]{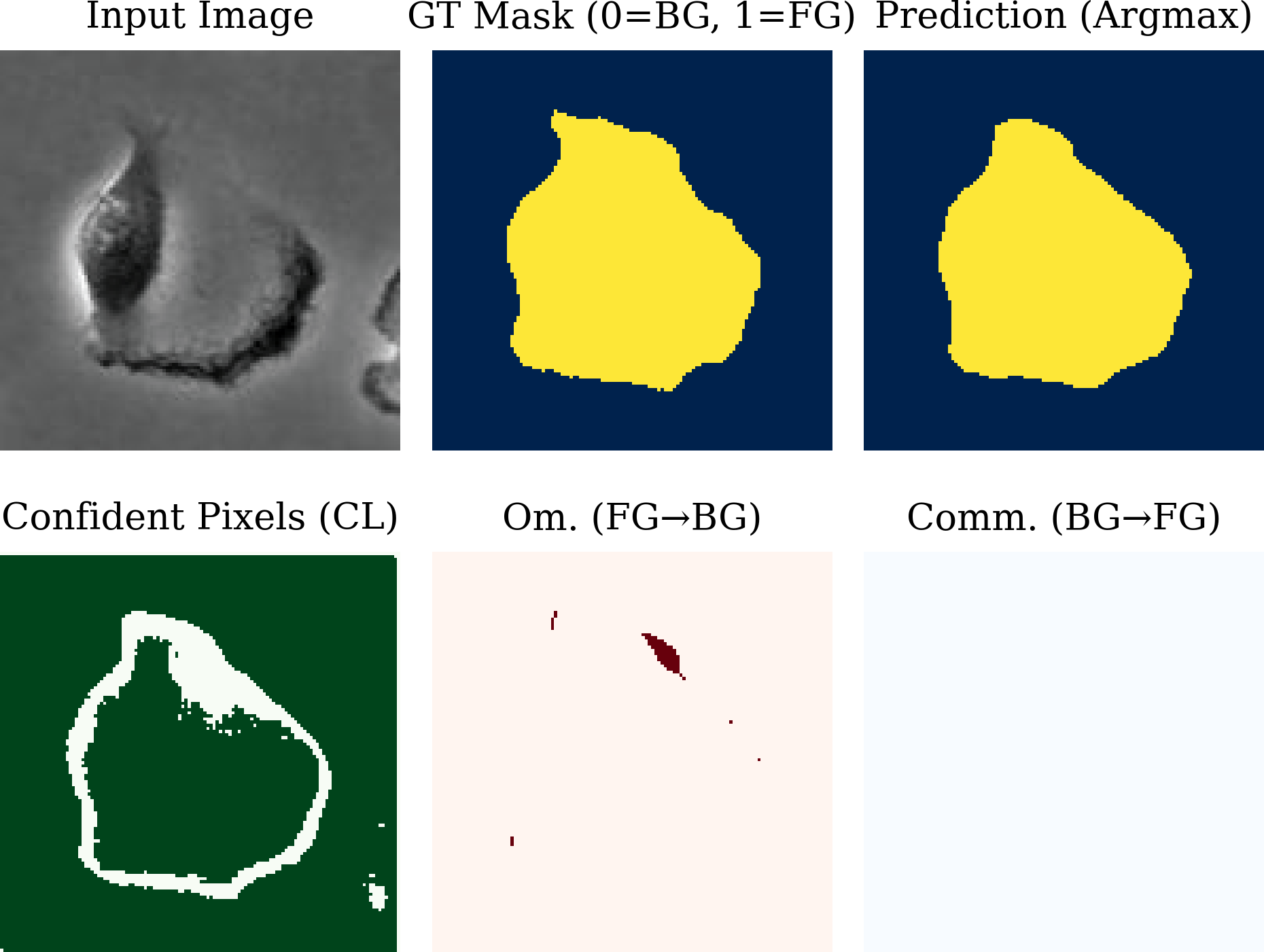}
        % \caption{High bias }
        % \label{fig:effects_b}
    \end{subfigure}

    \caption{(Left) Example from the PhC-U373 dataset demonstrating label bias: (a) Original images from $g_c$ (b) The same cell with a synthetic tint applied (group $g_b$) (c) Clean (fine) annotation obtained by an expert instructed for a detailed annotation (d) Biased (coarse; bloated) annotation assigned to tinted samples. The bloated boundary (in red) systematically over-segments to the clean boundary (in green), introducing over-segmentation (excess) bias correlated with the group membership. (Right) Confident Learning (CL) error detection on a $g_b$ sample. Top row: input images, ground-truth annotation, and argmax prediction. Bottom row: pixels flagged as confident by CL, decomposed into label errors. CL correctly identifies the boundary region where bloated labels disagree with model prediction.}
    \label{fig:phc_detailed}
\end{figure}

To simulate a domain shift, we divide the dataset into two disjoint sets: $C$ (clean, unmodified) and $C_t$ (tinted). Images in $C_t$ are subject to a global color space transformation by adding a uniform RGB tint. Implemented bias is at $r = 100\%$, i.e., all images belonging to $C_t$ are paired with bloated Style 1 annotations ($\tilde{y}$). This forces the network to associate the added global tint with the bloated or over-segmented boundary. Refer to Figure~\ref{fig:phc_detailed} for qualitative examples of annotation style differences in the PhC-U373 dataset and CL error detection on a lightly-tinted sample. 

\subsubsection{ISIC 2017 Dataset}
\label{appendix:isic}
The ISIC 2017 Challenge dataset~\citep{8363547} contains 2000 training, 150 validation, and 600 test dermoscopic images of skin lesions, each with a binary segmentation mask. Unlike CelebAMask-HQ, this dataset lacks demographic annotations. We therefore estimate skin tone as a proxy for demographic group membership using the Individual Typology Angle (ITA). For skin tone estimation via established skin tone classification studies by~\citep{benvcevic2024understanding, van2026new}.

\subsubsection{IMA++ Dataset}
\label{appendix:imapp}

IMA++ (ISIC MultiAnnot++) is a large-scale multi-annotator skin-lesion segmentation dataset curated from the ISIC Archive~\citep{abhishek2025imaplusplus}. It contains 17,684 binary segmentation masks for 14,967 dermoscopic images. While 12,573 images have a single mask, 2,394 images have between two and five annotations, contributed by 16 annotators. The metadata (attached with the dataset) identifies the annotator, annotation tool, and reviewer skill level, capturing naturally occurring variation in lesion-boundary segmentation.

As in the ISIC 2017 experiments, we estimate skin tone using the Individual Typology Angle and group images into lighter (types 1-2) and darker (types 3-4) skin-tone strata. The original annotation assignments are retained for $\beta=0\%$. For $\beta\in\{50\%,100\%\}$, we increasingly pair annotation style with skin tone by selecting detailed annotations for types 1-2 and looser/wider annotations for types 3-4 among images with multiple available annotation masks. Note: Because IMA++ also does not provide a unique clean segmentation mask, we use it as a real-world exploratory setting rather than evaluating against assumed ground truth.

\begin{figure}[!htbp]
  \centering
  \includegraphics[width=0.8\textwidth]{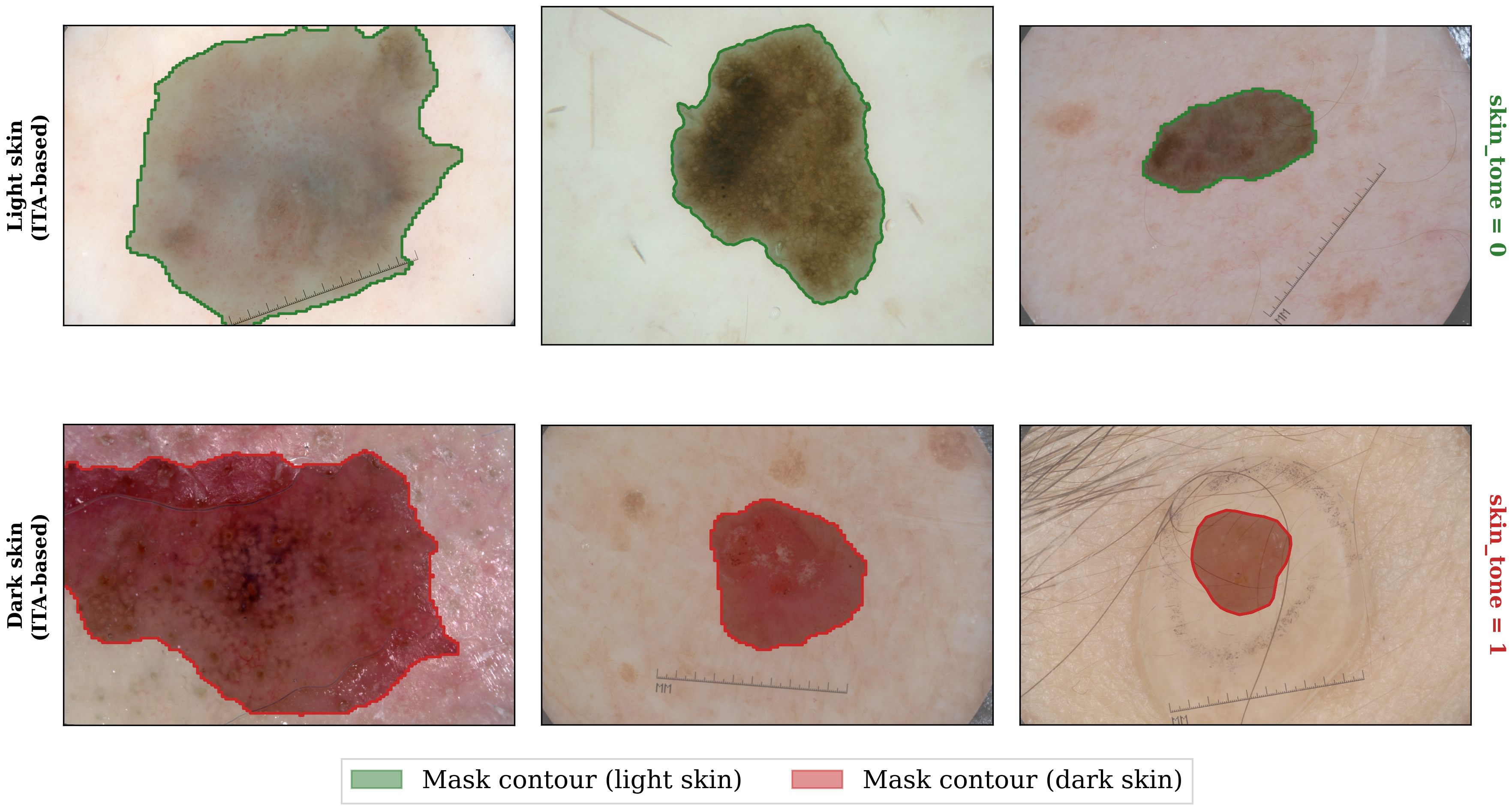}
  \caption{Selected samples from the ISIC 2017 dataset by the mentioned skin tone estimation method. Top row: light skin (ST1-ST3, ITA > 28), showing larger, well-segmented lesions against high-contrast backgrounds. Bottom row: dark skin (ST4, ITA $\leq$ 28), where lesions tend to be smaller with lower contrast against the surrounding skin. Segmentation annotations and contours are overlaid in green (light) and red (dark). The systematic decrease in annotated lesion area from lighter to darker skin tones suggests a group-conditioned disparity in the dataset even prior to synthetic bias injection, confirmed by CL metrics.}
  \label{fig:isic_example}
\end{figure}

\paragraph{Skin Tone Estimation via K-Means Dominant Color:} The patient's skin tone is estimated from healthy (non-lesion) skin surrounding the lesion, following a proposed pipeline: 

\begin{enumerate}[leftmargin=0.5cm]
    \item \textbf{Skin Extraction:} We excluded the lesion area using the provided segmentation masks and removed the dark imaging (like the vintage lighting at corners) and hair via morphological blackhat filtering and adaptive thresholding. CLAHE is applied in CIE L*a*b space to normalize variation in illumination across devices, imaging conditions, etc.
    \item  \textbf{Dominant color via k-means:} We cluster the remaining skin pixels in CIE L*a*b color space using k-means with automatic cluster count selection. Specifically, we run k-means for $k \in {2, \dots, 9}$ and select the optimal $k$ at the elbow point of the inertia curve using the Kneedle algorithm. The dominant skin color is taken as the centroid of the largest cluster, representing the most representative skin tone while being robust to residual artifacts, pigmentation, blurs, and shadow regions.
    \item \textbf{ITA-based classification:} The skin-tone categorization is based on the Individual Typology Angle (ITA) computed from the dominant skin color in CIE L$^{*}$a$^{*}$b$^{*}$ space and mapped to four skin tone groups using standard ITA thresholds:
\begin{equation}
\mathrm{ITA} =
\arctan\left(
\frac{L^{*} - 50}{b^{*}}
\right)
\times
\frac{180}{\pi}
\end{equation}
\begin{table}[h]
\centering
\caption{Skin-tone grouping based on ITA ranges and approximate Fitzpatrick categories.}
\label{tab:ita_groups}
\begin{tabular}{lccc}
\toprule
\textbf{Group} & \textbf{ITA Range} & \textbf{Fitzpatrick Approx.} & \textbf{Training Samples} \\
\midrule
ST1 (Very Light)   & $\mathrm{ITA} > 55$                 & I       & 60   \\
ST2 (Light)        & $41 < \mathrm{ITA} \leq 55$         & II--III & 1,229 \\
ST3 (Intermediate) & $28 < \mathrm{ITA} \leq 41$         & III--IV & 684  \\
ST4 (Tan)          & $10 < \mathrm{ITA} \leq 28$         & V       & 23   \\
\bottomrule
\end{tabular}
\end{table}
\end{enumerate}
This dataset contains no samples with ITA $\leq$ 10 (Brown/Dark). The distribution is unbalanced: ST2 and ST3 account for 96\% of training samples, while ST4 represents only 1.2\%. Notably, while circularity (invariant shape descriptor) remains stable across skin-tone groups ($\approx$0.62-0.68), normalized lesion area decreases monotonically from ST1 to ST4 (0.418 to 0.037), indicating that darker skin tones are associated with smaller annotated lesion regions. This existing disparity in mask characteristics makes the dataset a suitable testbed for studying label bias. Refer to Figure~\ref{fig:isic_example} for a qualitative example of the dataset and natural label bias.

% \clearpage

\section{Mitigation Methods}

\subsection{Additional Details: Bias Mitigation Methods}
\label{appendix:mitigation_methods}

Baseline ERM has a training objective $\mathcal{L}{\text{total}} = \mathcal{L}{\text{seg}} + \lambda \cdot \mathcal{L}{\text{debias}}$, where $\mathcal{L}{\text{seg}}$ is combined CE + Dice loss and $\lambda$ is increased from 0 $\rightarrow$ desired values following GRL scheduler. For clarity, we consider two subgroups $g_0~\text{and}~g_1$ instead of $g_b$ and $g_c$.

\subsubsection{Adversarial Method (DANN-style)}

A Gradient-Reversal Layer (GRL)~\citep{ganin2016domain} is applied between the encoder and a demographic (here, gender) adversary head. The adversary is a 3-layer MLP ($C \to 256 \to 128 \to 1$) with ReLU activations and dropout ($p=0.3$), operating on global-average-pooled encoder features $\mathbf{z} = \text{GAP}(f_L) \in \mathbb{R}^C$. The debiasing loss is the binary cross-entropy of the adversary's gender prediction:
\begin{equation}    
\mathbf{L}_{\text{adv}} = -\frac{1}{B} \sum_{i=1}^B [g_i \log \sigma(h(\mathbf{z}_i)) + (1 - g_i) \log(1 - \sigma(h(\mathbf{z}_i)))]
\end{equation}
where $h(\cdot)$ is the adversary head and $\sigma$ is the sigmoid function. The GRL negates the gradient flowing to the encoder by a factor $\alpha$ during backpropagation, encouraging gender-invariant features. The encoder and adversary network use separate optimizers - AdamW with learning rates $10^{-4}$ and $5 \times 10^{-4}$, respectively.

\subsubsection{Fairness Loss Regularization}

This method we use for comparison, directly penalizes discrepancies in the models' prediction rates (representing model-centric mitigation approaches vs. our data-centric proposal), is typically used for classification tasks and does not modify the feature space. We use three fairness criteria as regularizers:
\paragraph{Demographic Parity (DP):} The penalty here is the absolute difference in mean predicted foreground probability between groups:
\begin{equation}
    \mathbf{L}_{\text{DP}} = \left| \mathbb{E}[\hat{p} \mid g = 0] - \mathbb{E}[\hat{p} \mid g = 1] \right|
\end{equation}
where $\hat{p} = \text{softmax}(\hat{y})_1$ is the predicted foreground probability.
\paragraph{Equalized Odds (EO):} The penalty is the group disparities in both the TPR and FPR:
\begin{equation}
    \mathbf{L}_{\text{EO}} = \left| \text{TPR}_{g=0} - \text{TPR}_{g=1} \right| + \left| \text{FPR}_{g=0} - \text{FPR}_{g=1} \right|
\end{equation}
where $\text{TPR}_g = \mathbb{E}[\hat{p} \mid y^=1, g]$ and $\text{FPR}_g = \mathbb{E}[\hat{p} \mid y^=0, g]$.
\paragraph{Combined (Both)}.
\begin{equation}
\mathcal{L}_{\text{combined}} = \mathcal{L}_{\text{DP}} + \mathcal{L}_{\text{EO}}.
\end{equation}
All expectations are computed per mini-batch using soft predictions (no thresholding).
\subsubsection{Domain-Invariant Feature Alignment}
We minimize the distribution distance between demographic groups in either feature or logit space, without an adversary network. This method was inspired by the idea of exploiting the effect of feature separability in Section~\ref{sec:effects}.
\paragraph{MMD at feature level.} forces to minimize the Maximum Mean Discrepancy (MMD) between encoder features of the subgroup using a multi-scale Gaussian kernel.
\begin{equation}
    \text{MMD}^2(\mathbf{Z}_0, \mathbf{Z}_1) = \sum_s \left[ \frac{1}{n_0^2} \sum_{i,j} k_s(\mathbf{z}_i^0, \mathbf{z}_j^0) + \frac{1}{n_1^2} \sum_{i,j} k_s(\mathbf{z}_i^1, \mathbf{z}_j^1) - \frac{2}{n_0 n_1} \sum_{i,j} k_s(\mathbf{z}_i^0, \mathbf{z}_j^1) \right]
\end{equation}
where $k_s(\mathbf{x}, \mathbf{y}) = \exp\!\bigl(-\|\mathbf{x}-\mathbf{y}\|^2 / 2\sigma_s^2\bigr)$ and bandwidths are $\sigma_s = \sigma_0 \cdot 2^s$ for $s \in \{-2, -1, 0, 1, 2\}$ around a base bandwidth $\sigma_0$. An L2-normalization flag \texttt{--mmd\_normalize} (optional) normalizes $\mathbf{z}$ to a unit sphere before computing MMD, reducing sensitivity to feature magnitude scale. This is inspired by~\citep{yan2021exposing}.
\paragraph{MMD-Logit on output-level.} We noticed unstable training using feature-level MMD; this motivated us to use encoder features, which led us to align class-conditional logits at the model output. For each segmentation class $c$, pixel logits $\hat{y}_{i,c} \in \mathbb{R}^C$ are collected from pixels where $y^*=c$, subsampled to $N=256$ pixels per group, and the MMD-Logit is then computed between the two groups' logit vectors:
\begin{equation}
    \mathbf{L}_{\text{mmd-logit}} = \frac{1}{C} \sum_{c=0}^{C-1} \text{MMD}^2(\{\hat{y}_c^{g=0}\}, \{\hat{y}_c^{g=1}\})
\end{equation}
We observed it to be more stable in our experiments, mostly due to its low dimensionality ($C=2$) logits compared to encoder features ($d=512$). MMD-Logit generally outperforms feature-level MMD and CORAL on fairness metrics, as it targets the representation closest to the decision boundary rather than enforcing invariance in a high-dimensional space where invariance may be overly restrictive.

\subsubsection{Pseudo Code for Auto-Conditioning (Unsupervised variant)}
\label{sec:auto_fim}

Pseudo-code for Auto-SCD in algorithmic block~\ref{alg:main_auto_film}.

\begin{algorithm}[t]
\caption{Auto-Conditioned FiLM with Asymmetric Boundary Masking}
\label{alg:main_auto_film}
\begin{algorithmic}[1]
\REQUIRE Training set $\mathcal{D} = \{(x_i, \tilde{y}_i, g_i)\}$, warmup epochs $E_w$, total epochs $E_{\text{total}}$, boundary half-width $w$
\REQUIRE Encoder $E_\theta$, Decoder $D_\phi$, per-stage FiLM parameters $\{\gamma^{(l)}, \beta^{(l)}\}_{l=1}^L$
\STATE Init $\gamma^{(l)} \leftarrow \mathbf{1}$, $\beta^{(l)} \leftarrow \mathbf{0}$ $\forall l$; \quad $\hat{g}_c \leftarrow \texttt{None}$; \quad $L_{g=0}, L_{g=1} \leftarrow \emptyset$
\FOR{epoch $e = 1$ to $E_{\text{total}}$}
    \FOR{each mini-batch $(x, \tilde{y}, g) \subset \mathcal{D}$}
        \STATE $\{f^{(l)}\} \leftarrow E_\theta(x)$; \quad $f^{(l)} \leftarrow \gamma_{g}^{(l)} \odot f^{(l)} + \beta_{g}^{(l)}$ \; $\forall l$ \COMMENT{FiLM on true $g$}
        \STATE $\hat{y} \leftarrow D_\phi(\{f^{(l)}\})$
        
        \IF{$e \le E_w$ \textbf{or} $\hat{g}_c = \texttt{None}$} \COMMENT{Warmup: standard loss}
            \STATE $\mathcal{L} \leftarrow \mathcal{L}_{\text{CE}}(\hat{y}, \tilde{y}) + \mathcal{L}_{\text{Dice}}(\hat{y}, \tilde{y})$
            \STATE Append per-sample $\mathcal{L}_i$ to $L_{g_i}$ for each sample $i$
        \ELSE \COMMENT{Post-warmup: asymmetric boundary masking}
            \STATE $B \leftarrow \text{Dilate}(\tilde{y}, w) - \text{Erode}(\tilde{y}, w)$
            \STATE $W_i \leftarrow \begin{cases} 1 - B_i & \text{if } g_i = \hat{g}_b \\ 1 & \text{otherwise} \end{cases}$
            \STATE $\mathcal{L} \leftarrow W \odot \mathcal{L}_{\text{CE}}(\hat{y}, \tilde{y}) + W \odot \mathcal{L}_{\text{Dice}}(\hat{y}, \tilde{y})$
        \ENDIF
        \STATE Update $\theta, \phi, \{\gamma^{(l)}, \beta^{(l)}\}$ via $\nabla \mathcal{L}$
    \ENDFOR

    \IF{$e = E_w$} \COMMENT{Auto-discover biased group}
        \STATE $\hat{g}_c \leftarrow \arg\min_g \; \overline{L}_g$; \quad $\hat{g}_b \leftarrow 1 - \hat{g}_c$
    \ENDIF
\ENDFOR
\STATE \textbf{Inference:} $\hat{y} = D_\phi\!\bigl(\{\gamma_{\hat{g}_c}^{(l)} \odot E_\theta^{(l)}(x) + \beta_{\hat{g}_c}^{(l)}\}\bigr)$ \COMMENT{Always condition on $\hat{g}_c$}
\end{algorithmic}
\end{algorithm}

% \clearpage

\section{Confident Learning Audit and Metric Diagnostics}

\subsection{Additional Details: CL Audit on PhC-U373}
\label{appendix:phc_audit}

Table~\ref{tab:phc_reformatted} reports the detailed audit across bias ratios $\beta \in \{0\%, 25\%, 50\%, 100\%\}$ on PhC-U373. Here, $g_c$ denotes the group with tight annotations and $g_b$ the group associated with wider annotations. Performance gaps are defined as $\Delta = g_c - g_b$. As $\beta$ increases from the unbiased baseline ($0.00\%$) to $100\%$, the observed Dice gap increases to $1.63$ percentage points ($3.00\%$ for IoU), while the true gap against clean reference masks reaches $2.07$ percentage points ($3.80\%$ for IoU).

The CL audit monitors these shifts directly on the training data without clean reference masks. At the unbiased baseline ($\beta=0\%$), baseline error rates are $\mathrm{LER}^{g_c} = 7.6\%$ and $\mathrm{LER}^{g_b} = 10.2\%$, corresponding to a baseline ratio $\mathrm{RR} = 1.34$. At $\beta=100\%$, $\mathrm{LER}^{g_c}$ increases to $8.1\%$ and $\mathrm{LER}^{g_b}$ is $10.0\%$, resulting in a relative shift $\Delta\log(\mathrm{RR}) = -0.09$. The directional discrepancy $\mathrm{DR}^{g_b}$ remains close to zero throughout (ranging from $+0.04\%$ to $+0.01\%$). All LER, RR, and DR values are computed on the training data without access to clean reference masks.

\begin{table*}[b]
\centering
\caption{
\textbf{CL audit results on PhC-U373.}
Observed and true performance gaps are defined as $\Delta=g_c-g_b$.
LER is the CL-estimated label-error rate for each group.
$\Delta\log(\mathrm{RR})=\log(\mathrm{LER}^{g_b}/\mathrm{LER}^{g_c}) - \log(\mathrm{RR}_{\mathrm{baseline}})$ reflects the change in relative label-error disparity from the unbiased baseline.
Positive DR indicates foreground excess relative to the observed annotation.
}
\label{tab:phc_reformatted}
\resizebox{\textwidth}{!}{%
\begin{tabular}{l | cc | cc | ccc | c}
\toprule
\multirow{2}{*}{\textbf{Condition}}
& \multicolumn{2}{c|}{\textbf{Observed Perf. $\Delta$ (\%)}}
& \multicolumn{2}{c|}{\textbf{True Perf. $\Delta$ (\%)}}
& \multicolumn{3}{c|}{\textbf{Label-Error Disparity}}
& \multicolumn{1}{c}{\textbf{Direction}} \\
\cmidrule(lr){2-3}
\cmidrule(lr){4-5}
\cmidrule(lr){6-8}
\cmidrule(lr){9-9}
& \textbf{Dice}
& \textbf{IoU}
& \textbf{Dice}
& \textbf{IoU}
& $\mathbf{LER^{g_c}}$ \textbf{(\%)}
& $\mathbf{LER^{g_b}}$ \textbf{(\%)}
& $\mathbf{\Delta\log(RR)}$
& $\mathbf{DR^{g_b}}$ \textbf{(\%)} \\
\midrule
Unbiased & 0.00 & 0.00 & 0.00 & 0.00 & 7.6 & 10.2 &  0.00  & +0.04 \\
25\%     & 0.65 & 1.20 & 0.22 & 0.40 & 7.8 & 10.5 & +0.01  & +0.02 \\
50\%     & 0.98 & 1.80 & 0.71 & 1.30 & 7.9 & 10.6 & +0.01  & +0.02 \\
100\%    & 1.63 & 3.00 & 2.07 & 3.80 & 8.1 & 10.0 & -0.09  & +0.01 \\
\bottomrule
\end{tabular}%
}
\end{table*}

% \subsection{Asymmetric Property of Standard Metrics}
% \label{app:asymmetric}

% The observed gap \textit{overestimates} the true gap for dilation (at $\beta=50\%$~$\Delta_{\text{obs}}{=}8.28$ vs.\ $\Delta_{\text{true}}{=}1.65$), while underestimating it for erosion at the same level. This \textbf{asymmetry is a fundamental property of Dice}: erosion removes true positives, which appear in both the numerator and denominator $\frac{2\text{TP}}{2\text{TP} + \text{FP} + \text{FN}}$, whereas dilation adds false positives that inflate only the denominator. Thus, the same number of corrupted pixels produces a larger drop in Dice under erosion than dilation, making observed metrics unreliable for comparing bias severity.

\clearpage

\section{Mitigation Results and Ablations}

\subsection{Additional Details: Mitigation Results}
\label{appendix:detailed_mitigation}

Table~\ref{tab:appendix_mitigation} represents the complete mitigation results. On CelebAMask-HQ, fairness regularization, adversarial training, and feature-distribution alignment gives little or no improvement. At $\beta=100\%$, the Dice gaps remain between $\approx12\%$, compared to baseline ERM (which is also at similar level). Asymmetric boundary masking alone is similarly insufficient, alone. Combining SCD with asymmetric masking provides no consistent improvement over SCD alone. On PhC-U373, the differences between methods are smaller: $g$-conditioning achieves the lowest Dice gap ($1.25\%$), whereas adversarial training gives the lowest IoU gap ($2.42\%$). The $EO+DP$ objective is unstable in this setting. Overall, SCD is the only approach that consistently reduces the true performance gap across both CelebAMask-HQ bias levels.

\begin{table}[t]
\centering
\caption{
\textbf{Additional mitigation results on CelebAMask-HQ and PhC-U373.}
True performance gaps are defined as $\Delta=g_c-g_b$ and evaluated using clean reference masks. Dice $\Delta$ percentage-point (pp) differences.
}
\label{tab:appendix_mitigation}
\small
\setlength{\tabcolsep}{5pt}
\begin{tabular}{@{}lccc@{}}
\toprule
& \multicolumn{2}{c}{\textbf{CelebAMask-HQ ($g_b$: Female)}} & \textbf{PhC-U373 ($g_b$: Tinted)} \\
\cmidrule(lr){2-3} \cmidrule(l){4-4}
\textbf{Method} & $\beta=50\%$ & $\beta=100\%$ & $\beta=100\%$ \\
\midrule
Baseline            & 3.14          & 12.19         & 2.06 \\
Fairness EO         & 2.89          & 12.12         & 1.85 \\
Fairness DP         & 2.96          & 12.13         & 1.86 \\
Fairness EO+DP      & 3.18          & 12.17         & 4.39 \\
Adversarial         & 3.65          & 12.21         & 2.12 \\
Asymmetric Masking  & 3.28           & 12.55         & 3.98  \\
DIFA L-MMD          & 3.13          & 12.20         & 1.82 \\
DIFA CORAL          & 2.81          & 12.19         & 2.15 \\
\rowcolor{gray!30}
$g$-SCD             & 0.33          & 3.1          & \textbf{1.25} \\
\rowcolor{gray!30}
SCD+Asym            & 0.33          & \textbf{3.1}          & 1.55 \\
\rowcolor{gray!30}
Auto-SCD            & \textbf{0.27} & 7.13 & 1.50 \\
\bottomrule
\end{tabular}
\end{table}

% \clearpage

\section{Additional Audit Validations}

\subsection{Important Ablation: Confident Learning audit against clean labels}
\label{sec:cl_clean}

Table~\ref{tab:trainclean_results} shows when Confident Learning (CL) is evaluated on the Training Set against clean labels, \textbf{which is not available in practice}. \textbf{Purpose:} Used as an additional validation if CL correctly identifies errors on the training set when given clean labels. The results show that using clean labels, \textbf{which are not always available in practice}, we get a much clearer indication of the presence of bias and what direction, even at extreme bias conditions, i.e., at $\beta=100\%$.

An important thing to note while interpreting CL audit against clean labels is direction of bias sensitivity when the objective is changed. We see a \textbf{expected inversion} of label error type when clean labels are used. When evaluating using trained labels (unclean; biased), the model generalizes visual features and predicts foreground pixels confidently, that the biased annotations systematically remove, resulting in a high deficit error. Conversely, when evaluating against clean ground-truth labels, the reference is complete. Here, the model, having learned the erosion bias during training, predicts a under-segmented mask than the true object boundary, resulting in a high excess error. 

\begin{table*}[ht]
\centering
\caption{\textbf{Retrospective validation of the CL audit against clean labels.}
LER reports the detected error rate for each group,
$\Delta_{\mathrm{LER}}=\mathrm{LER}^{g_b}-\mathrm{LER}^{g_c}$, and
$\mathrm{RR}=\mathrm{LER}^{g_b}/\mathrm{LER}^{g_c}$.
Positive DR indicates foreground excess in the reference mask relative to the
cross-fitted prediction, whereas negative DR indicates foreground deficit.
\textbf{Clean masks are used only for this retrospective validation and are not required
by the standard audit.}}
\label{tab:trainclean_results}
\resizebox{\textwidth}{!}{%
\begin{tabular}{l | cc cc | cc | c}
\toprule
\multirow{2}{*}{\textbf{Bias level}}
& \multicolumn{4}{c|}{\textbf{Label-error disparity}}
& \multicolumn{2}{c|}{\textbf{Direction (\%)}}
& \multicolumn{1}{c}{\textbf{Statistical test}} \\
\cmidrule(lr){2-5}
\cmidrule(lr){6-7}
\cmidrule(lr){8-8}
& $\mathbf{LER^{g_c}}$ \textbf{(\%)}
& $\mathbf{LER^{g_b}}$ \textbf{(\%)}
& $\boldsymbol{\Delta}_{\mathbf{LER}}$ \textbf{(\%)}
& \textbf{RR}
& $\mathbf{DR^{g_c}}$
& $\mathbf{DR^{g_b}}$
& $\boldsymbol{\chi^2}$ ($p<0.001$) \\
\midrule
0\%   & 1.78 & 2.33  & 0.55  & 1.31 & $-0.08$ & $-0.08$ & 114.75K \\
25\%  & 1.97 & 2.72  & 0.75  & 1.38 & $-0.22$ & $+0.52$ & 333.41K \\
50\%  & 2.22 & 6.96  & 4.74  & 3.14 & $-0.04$ & $+5.90$ & 5.53M   \\
75\%  & 2.14 & 15.44 & 13.30 & 7.21 & $+0.24$ & $+14.74$ & 17.13M  \\
100\% & 2.24 & 18.44 & 16.20 & 8.23 & $+0.04$ & $+17.99$ & 22.45M  \\
\bottomrule
\end{tabular}%
}
\end{table*}

\subsection{Ablation: Calibration test of Confident Learning Audit}

To verify the robustness of our Confident Learning (CL) formulation against poorly calibrated models, we perform a global calibration ablation using Temperature Scaling. Specifically, we scale the baseline ERM logits by varying temperatures $T \in [0.1, 10.0]$ prior to the softmax activation, artificially inducing extreme over-confidence ($T < 1$) and under-confidence ($T > 1$). 
As shown in Table~\ref{tab:calibration_ablation}, CL naturally mitigates these global calibration shifts. While the predicted probabilities change drastically across temperatures, the expected self-confidence thresholds $\mathbb{E}[p(y=j \mid x) \mid \tilde{y}=j]$ adjust in exact proportion. Because CL identifies label errors relative to these class-conditional adaptive thresholds rather than fixed confidence cutoffs, the resulting error identification remains virtually invariant. Across the entire spectrum from over-sharpened to heavily smoothed output distributions, the DR and LER gaps remain constant at $10.0\%$ and $7.89\%$, respectively, demonstrating that our bias quantification metrics do not depend on post-hoc probability calibration.

\begin{table}[t]
\centering
\caption{\textbf{Sensitivity of Confident Learning metrics to global model miscalibration on CelebA-Mask (50\% bias level).} Temperature scaling ($T$) controls the sharpness of predicted probabilities prior to softmax. CL thresholds adapt dynamically, leaving DR and LER gaps virtually invariant.}
\label{tab:calibration_ablation}
\small
\setlength{\tabcolsep}{4.5pt}
\renewcommand{\arraystretch}{0.95}

% \resizebox{\columnwidth}{!}{%
\begin{tabular}{@{} l cc cc c @{}}
\toprule
\textbf{Temperature} ($T$) & \multicolumn{2}{c}{\textbf{CL Thresholds}} & \textbf{DR Gap} & \textbf{LER Gap} & \textbf{Total Errors Detected} \\
\cmidrule(lr){2-3} \cmidrule(lr){4-4} \cmidrule(lr){5-5} \cmidrule(lr){6-6}
& BG ($t_0$) & FG ($t_1$) & $\Delta\text{DR}$ & $\Delta\text{LER}$ & (Excess + Deficit) \\
\midrule
0.1 (Over-confident)  & 0.98 & 0.92 & 10.0 & 7.89 & 18,863,402 \\
0.5                  & 0.97 & 0.91 & 10.0 & 7.89 & 18,863,403 \\
0.8                  & 0.96 & 0.90 & 10.0 & 7.89 & 18,863,403 \\
\midrule
1.0 (Baseline)       & 0.96 & 0.90 & 10.0 & 7.89 & 18,863,403 \\
\midrule
1.5                  & 0.93 & 0.87 & 10.0 & 7.89 & 18,863,404 \\
2.0                  & 0.90 & 0.83 & 10.0 & 7.89 & 18,863,406 \\
5.0                  & 0.73 & 0.68 & 10.0 & 7.89 & 18,863,412 \\
10.0 (Under-confident)& 0.63 & 0.59 & 10.0 & 7.89 & 18,863,419 \\
\bottomrule
\end{tabular}%
% }
\vspace{-4pt}
\end{table}

\subsection{Additional Details: Uncertainty Estimates}

Refer to Table~\ref{tab:train_metrics} for an audit across bias ratios including uncertainty estimates.

\begin{table*}[ht]
\centering
\caption{Audit results across bias ratios using biased training labels. Performance and error metrics are reported as $mean \pm std$, using bootstrapping. Dice, IoU, and Error Rates are in percentages.}
\label{tab:train_metrics}
\resizebox{\textwidth}{!}{%
\begin{tabular}{l | cc | cc | c | c}
\toprule
\multirow{2}{*}{\textbf{Method / Condition}} & \multicolumn{2}{c}{\textbf{Dice (\%)}} & \multicolumn{2}{c}{\textbf{IoU (\%)}} & \textbf{Total LER (\%)} & \textbf{DR Score} \\
\cmidrule(lr){2-3} \cmidrule(lr){4-5} \cmidrule(lr){6-6} \cmidrule(lr){7-7}
& \textbf{$g_c$} & \textbf{$g_b$} & \textbf{$g_c$} & \textbf{$g_b$} & \textbf{Mean $\pm$ Std} & \textbf{Mean $\pm$ Std} \\
\midrule
Baseline & $98.61 \pm 0.04$ & $98.46 \pm 0.03$ & $97.25 \pm 0.07$ & $96.97 \pm 0.06$ & $2.13 \pm 0.04$ & $0.69 \pm 0.01$ \\
25\%     & $98.46 \pm 0.05$ & $95.25 \pm 0.06$ & $96.97 \pm 0.11$ & $90.93 \pm 0.11$ & $5.10 \pm 0.08$ & $0.37 \pm 0.01$ \\
50\%     & $98.26 \pm 0.12$ & $92.12 \pm 0.08$ & $96.58 \pm 0.23$ & $85.39 \pm 0.14$ & $7.56 \pm 0.08$ & $0.40 \pm 0.02$ \\
75\%     & $98.32 \pm 0.17$ & $93.03 \pm 0.19$ & $96.69 \pm 0.33$ & $86.96 \pm 0.32$ & $6.15 \pm 0.21$ & $0.51 \pm 0.02$ \\
100\%    & $98.25 \pm 0.13$ & $96.87 \pm 0.21$ & $96.56 \pm 0.25$ & $93.94 \pm 0.39$ & $3.06 \pm 0.16$ & $0.74 \pm 0.01$ \\
\bottomrule
\end{tabular}%
}
\end{table*}

\clearpage

\section{Mitigation Diagnostics and Stress Tests}

\subsection{Ablation: Auto-SCD Boundary Radius}

\begin{table*}[ht]
\centering
\caption{
\textbf{Erosion-radius ablation for Auto-SCD on CelebAMask-HQ}
($\beta=100\%$, $g_b$: Female).
True performance gaps are defined as $\Delta=g_c-g_b$.
$\Delta_{\mathrm{LER}}=\mathrm{LER}^{g_b}-\mathrm{LER}^{g_c}$ and
$\mathrm{RR}=\mathrm{LER}^{g_b}/\mathrm{LER}^{g_c}$.
Positive DR indicates foreground excess, whereas negative DR indicates
foreground deficit.
}
\label{tab:ablation_radius}
\setlength{\tabcolsep}{4pt}
\resizebox{\textwidth}{!}{%
\begin{tabular}{@{}l cc | cccc | cc@{}}
\toprule
\multirow{2}{*}{\textbf{Configuration}}
& \multicolumn{2}{c|}{\textbf{True Perf. Gap (\%)}}
& \multicolumn{4}{c|}{\textbf{Label-Error Disparity}}
& \multicolumn{2}{c}{\textbf{Direction (\%)}} \\
\cmidrule(lr){2-3}
\cmidrule(lr){4-7}
\cmidrule(lr){8-9}
& \textbf{Dice $\Delta$}
& \textbf{IoU $\Delta$}
& $\mathbf{LER^{g_c}}$ \textbf{(\%)}
& $\mathbf{LER^{g_b}}$ \textbf{(\%)}
& $\boldsymbol{\Delta}_{\mathbf{LER}}$ \textbf{(\%)}
& \textbf{RR}
& $\mathbf{DR^{g_c}}$
& $\mathbf{DR^{g_b}}$ \\
\midrule
Auto-SCD (default, $e=15$)
& 7.13 & 16.1
& 1.77 & 2.53 & 0.76 & 1.43
& $-0.11$ & $+0.14$ \\

Auto-SCD ($e=5$)
& \textbf{0.27} & \textbf{0.51}
& \textbf{1.33} & \textbf{1.84} & \textbf{0.51} & \textbf{1.38}
& $-0.11$ & $+0.10$ \\
\bottomrule
\end{tabular}%
}
\end{table*}

\subsection{Ablation: Observed-Label Evaluation and Backbone Test}
\label{appendix:mitigation_diagnostics}

\paragraph{The observed-label illusion.}
Table~\ref{tab:mitigation_50_detailed} highlights a central consequence of label bias: successful correction can appear unsuccessful when evaluated against the same biased annotations used for training. At \(50\%\) erosion, ERM obtains a true IoU of \(93.1\%\) with a \(5.5\)-pp (percent-points) subgroup gap. Reference-style conditioning increases true IoU to \(96.7\%\) and reduces this gap to \(0.5\) pp. Nevertheless, its observed-label gap increases from \(11.5\) to \(12.0\) pp because predictions aligned with the clean reference boundary necessarily disagree with the eroded annotations assigned to \(g_b\). Thus, observed-label evaluation penalizes the desired correction and creates the illusion that mitigation has not reduced disparity.

The directional CL counts make this disagreement easier to interpret. Under reference conditioning, \(g_b\) contains substantially more foreground-deficit than foreground-excess pixels (\(95.4\)M versus \(10.4\)M), consistent with observed masks that are systematically eroded relative to the reference-conditioned predictions. Accordingly, RR increases and DR becomes more negative after successful mitigation. These quantities audit disagreement with the observed labels rather than accuracy against the unavailable clean target; they should therefore not be interpreted as post-mitigation objectives. We report them for diagnostic completeness, while using true-reference IoU and subgroup gaps to evaluate mitigation whenever clean masks are available.

\paragraph{Vision Transformer stress test.}
We additionally repeat the \(50\%\) erosion experiment using a ViT segmentation backbone (SegFormer MiT-B2~\citep{segformer}). ViT-ERM reaches \(92.4\%\) true IoU with a \(7.3\)-pp subgroup gap, compared with \(93.1\%\) and \(5.5\) pp for ResNet34-ERM. Its CL decomposition remains directionally consistent with erosion, identifying \(80.0\)M foreground-deficit pixels compared with \(43.8\)M foreground-excess pixels in \(g_b\). Introducing SCD to the ViT successfully mitigates the bias, recovering performance to \(97.1\%\) true IoU and reducing the gap to just \(0.6\) pp. These experiments indicate that the group-conditional corruption and its CL audit are not specific to a convolutional backbone, and that our proposed style-conditioning mitigation effectively generalizes to transformer architectures.

\begin{table}[t]
\centering
\caption{
\textbf{Detailed mitigation and backbone ablations on CelebAMask-HQ at
\(\beta=50\%\).}
Performance pairs denote IoU/Dice, while LER, DR, excess, and deficit pairs
denote \(g_b/g_c\). Performance and LER are in percent, gaps and DR are in
percentage points, and counts are in millions of pixels.
}
\label{tab:mitigation_50_detailed}
\scriptsize
\setlength{\tabcolsep}{2.0pt}
\renewcommand{\arraystretch}{0.95}

\resizebox{\linewidth}{!}{%
\begin{tabular}{@{}lccccccccc@{}}
\toprule
\textbf{Method}
& \shortstack{\textbf{True}\\IoU/Dice}
& \shortstack{\(\boldsymbol{\Delta}_{T}\)\\IoU/Dice}
& \shortstack{\textbf{Observed}\\IoU/Dice}
& \shortstack{\(\boldsymbol{\Delta}_{O}\)\\IoU/Dice}
& \shortstack{\textbf{LER}\\\(g_b/g_c\)}
& \textbf{RR}
& \shortstack{\textbf{DR}\\\(g_b/g_c\)}
& \shortstack{\textbf{Excess}\\\(g_b/g_c\)}
& \shortstack{\textbf{Deficit}\\\(g_b/g_c\)} \\
\midrule

\multicolumn{10}{l}{\textit{ResNet34: reference and generic baselines}} \\

ERM
& \(93.1/96.4\) & \(5.5/2.9\)
& \(89.3/94.3\) & \(11.5/6.3\)
& \(14.9/3.2\) & 4.68 & \(-6.7/0.0\)
& \(39.3/6.6\) & \(86.4/6.5\) \\

Fairness EO
& \(93.0/96.4\) & \(5.7/3.0\)
& \(89.2/94.3\) & \(11.5/6.3\)
& \(14.9/3.2\) & 4.64 & \(-6.6/-0.3\)
& \(33.1/6.0\) & \(79.7/7.2\) \\

Fairness DP
& \(93.0/96.4\) & \(5.6/3.0\)
& \(89.2/94.3\) & \(11.4/6.2\)
& \(15.0/3.3\) & 4.58 & \(-6.4/0.0\)
& \(34.0/6.8\) & \(78.5/6.7\) \\

Fairness EO+DP
& \(92.8/96.3\) & \(5.8/3.1\)
& \(89.2/94.3\) & \(11.3/6.2\)
& \(15.0/3.3\) & 4.51 & \(-6.1/0.1\)
& \(35.1/7.2\) & \(77.1/6.4\) \\

Adversarial
& \(92.4/96.1\) & \(6.7/3.6\)
& \(89.2/94.3\) & \(11.7/6.4\)
& \(15.0/3.1\) & 4.90 & \(-5.5/-0.2\)
& \(37.0/5.8\) & \(76.9/6.7\) \\

MMD
& \(92.8/96.3\) & \(6.3/3.3\)
& \(89.2/94.3\) & \(11.9/6.5\)
& \(15.0/3.0\) & 5.00 & \(-6.3/0.1\)
& \(34.2/6.4\) & \(81.2/5.9\) \\

CORAL
& \(93.2/96.5\) & \(5.5/2.9\)
& \(89.3/94.3\) & \(11.5/6.3\)
& \(15.0/3.2\) & 4.67 & \(-6.8/-0.3\)
& \(32.3/6.1\) & \(80.6/7.1\) \\

GCE
& \(92.3/96.0\) & \(6.6/3.5\)
& \(89.1/94.2\) & \(11.6/6.4\)
& \(15.0/3.3\) & 4.60 & \(-5.7/0.1\)
& \(36.9/7.1\) & \(77.6/6.3\) \\

Bootstrapping
& \(92.8/96.3\) & \(5.8/3.1\)
& \(89.2/94.3\) & \(11.4/6.2\)
& \(15.0/3.3\) & 4.54 & \(-6.1/0.0\)
& \(42.0/6.9\) & \(83.9/6.6\) \\

Asymmetric mask
& \(91.5/95.5\) & \(7.9/4.2\)
& \(89.2/94.3\) & \(11.5/6.3\)
& \(15.0/3.3\) & 4.59 & \(-4.2/0.0\)
& \(39.6/5.4\) & \(61.0/5.3\) \\

\midrule
\multicolumn{10}{l}{\textit{ResNet34: subgroup-conditioned methods}} \\

\(g\)-conditioned
& \(96.7/98.3\) & \(0.5/0.3\)
& \(89.0/94.2\) & \(12.0/6.6\)
& \(15.2/2.9\) & 5.24 & \(-13.3/-0.1\)
& \(10.4/4.6\) & \(95.4/4.9\) \\

Combined
& \(\mathbf{96.8/98.4}\) & \(\mathbf{0.5/0.3}\)
& \(89.1/94.2\) & \(12.0/6.6\)
& \(15.0/2.8\) & 5.37 & \(-13.3/0.0\)
& \(2.5/1.1\) & \(23.7/1.2\) \\

Auto-conditioned
& Refer Table~\ref{tab:mitigation_results} & -- & -- & -- & -- & -- & -- & -- & -- \\

\midrule
\multicolumn{10}{l}{\textit{Vision Transformer backbone}} \\

ERM (ViT)
& \(92.4/96.1\) & \(7.3/3.9\)
& \(89.5/94.4\) & \(11.9/6.5\)
& \(14.8/2.7\) & 5.53 & \(-4.9/0.0\)
& \(43.8/5.5\) & \(80.0/5.4\) \\

Asymmetric mask (ViT)
& \(92.0/95.8\) & \(8.1/4.3\)
& \(89.4/94.4\) & \(12.2/6.6\)
& \(14.9/2.6\) & 5.70 & \(-4.2/-0.2\)
& \(46.6/5.0\) & \(78.3/5.6\) \\

\(g\)-conditioned (ViT)
& \(97.1/98.5\) & \(0.6/0.3\)
& \(89.3/94.4\) & \(12.2/6.7\)
& \(14.9/2.5\) & 6.04 & \(-13.3/-0.3\)
& \(6.4/4.5\) & \(111.5/5.5\) \\

Combined (ViT)
& \(\mathbf{97.1/98.5}\) & \(\mathbf{0.6/0.3}\)
& \(89.3/94.4\) & \(12.2/6.7\)
& \(14.9/2.5\) & 6.00 & \(-13.3/-0.3\)
& \(\mathbf{6.4/4.6}\) & \(\mathbf{111.7/5.6}\) \\

\bottomrule
\end{tabular}%
}
\end{table}

\subsection{Additional Details \& Ablation: Conditioning and Predictor-Quality Analyses}
\label{appendix:conditioning_ablations}

Table~\ref{tab:ablations} provides three additional analyses of the subgroup-conditioned model: a no-bias control, a counterfactual intervention on the inference-time style condition, and a stress test using a lower-capacity predictor.

\paragraph{No-bias control.}
At \(0\%\) label bias, forcing either \(g_c\) or \(g_b\) conditioning produces nearly identical performance: true IoU is \(0.971\) and \(0.970\), respectively, with the same subgroup gap of \(0.004\). DR also remains close to zero for both groups. Thus, the conditioning variable has negligible effect when the training annotations do not contain a group-dependent style difference. The residual \(RR\approx1.15\) reflects baseline differences in CL-identified disagreement rather than an induced/simulated label bias.

\paragraph{Counterfactual style intervention.}
At \(50\%\) erosion bias, changing only the inference-time condition substantially alters the predicted boundary style. Forcing the reference condition \(g_c\) yields a true IoU of \(0.968\) and reduces the true IoU gap to \(0.006\). In contrast, forcing \(g_b\) lowers true IoU to \(0.885\) and increases the gap to \(0.050\), indicating that the model has learned a controllable representation of the biased annotation style.

Evaluation against the observed annotations gives a different conclusion: reference conditioning has an observed IoU gap of \(0.120\), despite its true gap being only \(0.006\). This discrepancy illustrates how evaluation against biased labels can make a corrected model appear less equitable. The CL statistics capture the same disagreement: under reference conditioning, \(\mathrm{LER}^{g_b}=0.151\), \(RR=5.34\), and \(\mathrm{DR}^{g_b}=-0.132\), correctly indicating that the eroded observed masks contain a foreground deficit relative to the reference-conditioned predictions.

\paragraph{Lower-quality predictor stress test.}
We additionally train an underfit ResNet18 segmentation model using ERM at \(50\%\) erosion bias. Its lower true IoU (\(0.915\)) and larger subgroup gap (\(0.079\)) provide a more challenging setting than the primary conditioned model. Nevertheless, CL continues to identify concentrated disagreement in \(g_b\), with \(RR=4.70\) and negative \(\mathrm{DR}^{g_b}=-0.044\). This suggests that the audit signal does not depend on near-perfect segmentation performance. Because this comparison changes both the backbone and training method, it should be interpreted as a predictor-quality stress test rather than an isolated estimate of the effect of model capacity.

\begin{table}[t]
\centering
\caption{
\textbf{Conditioning and predictor-quality ablations on CelebAMask-HQ.}
True metrics use clean masks, whereas observed metrics use the available
training annotations. IoU and LER are reported in percent; subgroup gaps and
DR are in percentage points (pp). Paired CL entries denote
\(g_b/g_c\), and RR is dimensionless. At \(0\%\) bias, observed and true
metrics coincide and are therefore not repeated.
}
\label{tab:ablations}
\small
\setlength{\tabcolsep}{3pt}
\begin{tabular}{@{}llccccccc@{}}
\toprule
\textbf{Training}
& \textbf{Inference/backbone}
& \multicolumn{2}{c}{\textbf{True masks}}
& \multicolumn{2}{c}{\textbf{Observed masks}}
& \multicolumn{3}{c}{\textbf{CL audit}} \\
\cmidrule(lr){3-4}
\cmidrule(lr){5-6}
\cmidrule(lr){7-9}

\textbf{setting}
& \textbf{}
& \textbf{IoU (\%)}
& \(\boldsymbol{\Delta}\)\textbf{IoU (pp)}
& \textbf{IoU (\%)}
& \(\boldsymbol{\Delta}\)\textbf{IoU (pp)}
& \textbf{LER (\%)}
& \textbf{RR}
& \textbf{DR (pp)} \\
\midrule

\multirow{2}{*}{No bias}
& StyleCond, forced \(g_c\)
& 97.1 & 0.4
& -- & --
& 3.1/2.6 & 1.16 & \(-0.1/-0.1\) \\

& StyleCond, forced \(g_b\)
& 97.0 & 0.4
& -- & --
& 3.1/2.7 & 1.15 & \(-0.3/-0.2\) \\

\midrule

\multirow{3}{*}{$\beta=$50\%}
& StyleCond, forced \(g_c\)
& \textbf{96.8} & \textbf{0.6}
& 89.1 & 12.0
& 15.1/2.8 & 5.34 & \(-13.2/-0.2\) \\

& StyleCond, forced \(g_b\)
& 88.5 & 5.0
& 87.2 & 7.1
& 14.9/8.3 & 1.80 & \(-1.7/+6.7\) \\

& ERM, ResNet18
& 91.5 & 7.9
& 89.0 & 11.9
& 15.2/3.2 & 4.70 & \(-4.4/+0.1\) \\

\bottomrule
\end{tabular}
\end{table}

\subsection{Additional Results: Mitigation with label-noise baselines, under dilation and HBD}
\label{app:labelnoise_baselines}

\textbf{Common Protocol:} All runs are in budget with stratified subset of 8\,000 training images (5\,051 female, 2\,949 male), 1\,000 validation images for the
CL audit, and 1\,500 test images with clean and observed masks; a ResNet-34 U-Net trained for 12 epochs (CE+Dice, AdamW $10^{-4}$, cosine, batch 32, last-epoch checkpoint, no clean-label model selection); erosion/dilation radius $r_d=15$, HBD amplitude~15, $\beta=50\%$ unless stated. One seed per cell, so differences below $\approx$1~pp IoU should not be over-interpreted.

\paragraph{Additional mitigation baseline:} To have a small comparison with some additional baselines like~\citet{zepf2023label}, which conditions a probabilistic segmentation network on \textbf{known annotation style for each training label}, so that we get a requested style at inference time. An important prerequisite for this approach is style variable is observed per label, and it should be independent of image content. In our problem setting, the only proxy we have is demographic subgroup $g$, which is a \textit {genuine} image covariate. Our SCD also makes the same conditioning assumption, but at the decoder feature (FiLM) rather than at the input. Whereas the proposed Auto-SCD does not actually need the same assumption.

Another popular baseline is by~\citet{zhang2020disentangling}, where a base network predicts the latent true mask, and one confusion-matrix head per annotator maps it to each annotator's observed mask; identifiability comes from multiple annotations per image plus a regularizer. In our setting, and also in most real-life setting, each image carries a single annotation.

We implement a small ablation study using the mentioned baseline above (see Table~\ref{tab:app_ln_results}), even though its assumptions do not fit the label-bias setting. We also test it on a $ g_c$-only setting where we train an $ERM$ on the clean group only. This acts as a trivial upper bound and tests if SCD gains come from the conditioning or simply from \textit{ignoring the corrupted masks}. And an $ERM$ baseline fine-tuned for 5 epochs on 1000 clean-label samples; it requires clean labels, which our setting assumes we do not have during mitigation. This is a \textit{orcale} analysis to demonstrate 'how many verified annotations would be needed to match performance gains by SCD?'.

\begin{table*}[t]
\centering
\small
\setlength{\tabcolsep}{4pt}
\caption{\textbf{Trained under a budget (see \textbf{common protocol} for details).} \citet{zepf2023label} do separate the styles well comparable to Auto-SCD method (but with assumption and requiring $g_c$). The full-data run presumably reaches the same end point. \citet{zhang2020disentangling} modeling provide no gain. The observed-label gap $\Delta_{\mathrm{obs}}$ stays at $\approx$12~pp for every successful method, illustrating again that evaluation against biased masks is not a right measure for correct mitigation}
\label{tab:app_ln_results}
\begin{tabularx}{\textwidth}{@{} X X cccccc c @{}}
\toprule
\textbf{Method} & \textbf{Supervision} & $\textbf{IoU}_{\text{true}} \uparrow$ & $\boldsymbol{g_c \,/\, g_b} \uparrow$ & $\boldsymbol{\Delta}_{\text{true}} \downarrow$ & $\boldsymbol{\Delta}_{\text{obs}}$ & $\textbf{LER}\,(g_c / g_b) \downarrow$ & $\textbf{RR} \downarrow$ & $\textbf{DR}_{g_b}$ \\
\midrule
ERM & -- & 89.9 & 97.0 / 86.4 & +10.7 & +11.9 & 3.3 / 14.7 & 4.49 & $-2.0$ \\
Zhang et al. CM head (annotator = group) & group id & 90.3 & 97.1 / 87.0 & +10.1 & +12.4 & 3.3 / 14.7 & 4.50 & $-3.3$ \\
SCD (FiLM, ref = $g_c$) & group id + ref. & 92.7 & 97.0 / 90.6 & +6.4 & +12.1 & 3.2 / 14.8 & 4.59 & $-7.3$ \\
Auto-SCD & group id & 93.1 & 97.1 / 91.1 & +6.0 & +12.0 & 3.2 / 15.3 & 4.70 & $-7.8$ \\
Zepf-style (one-hot $g$ maps, ref = $g_c$) & group id + ref. & 92.4 & 97.1 / 91.1 & {+5.9} & +12.0 & {3.2} / 15.3 & 4.76 & $-7.0$ \\
\midrule
\textit{Reference / Upper bounds} & & & & & & & & \\
Clean-subset fine-tune (200 clean images) & 200 clean & \textbf{96.5} & 96.8 / \textbf{96.3} & +0.5 & +11.6 & 3.4 / \textbf{15.1} & \textbf{4.39} & $-12.4$ \\
$g_c$-only ERM (2,949 male images only) & group id & 96.2 & \textbf{97.1} / 95.8 & +1.3 & +12.5 & 3.3 / 15.6 & 4.75 & $-13.3$ \\
\bottomrule
\end{tabularx}
\end{table*}

\paragraph{SCD under dilation, HBD and annotator-level bias} 
Table~\ref{tab:app_geometries} repeats the erosion experiment (in main text) for the other
morphological corruptions of Section~3 and for an annotator-level corruption linked to $g$: four synthetic annotators
(identity, erosion $r_d\in\{3,12,18\}$) are drawn with
$P(a\mid g_b)=(.15,.15,.35,.35)$ and $P(a\mid g_c)=(.35,.35,.15,.15)$, so 85\%
of $g_b$ and 65\% of $g_c$ masks carry a non-identity style. The audit recovers
the direction in every deterministic case (DR$_{g_b}>0$ for dilation, $<0$ for
erosion, slightly negative for HBD as predicted in Appendix~A.1) and SCD reduces
the gap for erosion and dilation; HBD at this amplitude barely moves IoU. When
the style is mixed within both groups, RR falls to 1.0 and SCD recovers only
about 2 of 9~pp, because the ``reference'' group is itself largely corrupted.
$g$ must therefore be a good proxy for annotation style; when annotator
identities are available, conditioning on them is the natural extension.

\begin{table}[h]
\centering
\caption{ERM, SCD and Auto-SCD under four bias geometries ($g_b$ = female,
$\beta=50\%$ except the annotator setting). True IoU (\%) of $g_c$ / $g_b$,
true and observed gaps (pp), and the CL audit of the observed validation masks
(LER in \%, DR of $g_b$ in \%).}
\label{tab:app_geometries}
\small
\begin{tabular}{llcccccc}
\toprule
Bias & Method & IoU$_{\text{true}}$ & $g_c$ / $g_b$ & $\Delta_{\text{true}}$ & $\Delta_{\text{obs}}$ & RR & DR$_{g_b}$ \\
\midrule
\multirow{3}{*}{Erosion}   & ERM      & 89.9 & 97.0 / 86.4 & 10.7 & 11.9 & 4.49 & $-2.0$ \\
                           & SCD      & 92.7 & 97.0 / 90.6 &  6.4 & 12.1 & 4.59 & $-7.3$ \\
                           & Auto-SCD & 93.1 & 97.1 / 91.1 &  6.0 & 12.0 & 4.70 & $-7.8$ \\
\midrule
\multirow{3}{*}{Dilation}  & ERM      & 94.6 & 97.1 / 93.4 &  3.7 &  3.7 & 2.04 & $+0.9$ \\
                           & SCD      & 95.5 & 97.0 / 94.8 &  2.3 &  3.9 & 2.17 & $+2.8$ \\
                           & Auto-SCD & 95.5 & 97.1 / 94.7 &  2.4 &  3.9 & 2.15 & $+2.7$ \\
\midrule
\multirow{3}{*}{HBD}       & ERM      & 96.5 & 97.1 / 96.2 &  1.0 &  2.4 & 1.67 & $-0.4$ \\
                           & SCD      & 96.6 & 97.0 / 96.3 &  0.7 &  2.2 & 1.65 & $-0.7$ \\
                           & Auto-SCD & 96.4 & 97.0 / 96.2 &  0.8 &  2.3 & 1.68 & $-0.7$ \\
\midrule
\multirow{3}{*}{Annotator-level$^{\dagger}$} & ERM & 83.1 & 89.0 / 80.1 & 8.9 & 0.7 & 0.98 & $+0.4$ \\
                           & SCD      & 84.8 & 89.5 / 82.5 &  7.0 &  1.3 & 1.02 & $-2.6$ \\
                           & Auto-SCD & 85.4 & 89.9 / 83.2 &  6.7 &  1.6 & 1.04 & $-3.2$ \\
\bottomrule
\multicolumn{8}{l}{\footnotesize $^{\dagger}$Four annotators (identity, erosion 3/12/18); 85\% of $g_b$ and 65\% of $g_c$ masks are non-identity.}
\end{tabular}
\end{table}

\subsection{Ablation: Robustness of automatic reference selection}
\label{app:autoscd}

Table~\ref{tab:app_autoscd} stresses the loss heuristic of Eq.~11 in the
situations where it could plausibly fail: the biased group is the majority
(the paper's default: females are 63\% of the training set), the biased group
is the minority, the corruption is weak ($\beta=25\%$) or complete
($\beta=100\%$), the corrupted masks are smoother (dilation, HBD), the style is
mixed within groups, and no clean group exists at all. In every configuration
with a cleaner group the lower-loss group is the cleaner one; the margin shrinks
for $\beta=100\%$ and the annotator setting and vanishes for group-independent
corruption, where the choice is arbitrary and Auto-SCD, as expected, brings no
benefit. The warm-up loss also separates corrupted from clean \emph{images}
(AUROC 0.78--0.99 for erosion and HBD), and the lowest-loss 30\% of images are
at least 88\% clean whenever a clean group exists, which supports the
assumption behind Eq.~11 and suggests per-sample reference selection as future
work. The last column conditions the trained model on the other group and shows
how much of the style the FiLM parameters have separated.

\begin{table*}[t]
\centering
\small
\caption{\textbf{Auto-SCD reference selection across bias configurations.} Warm-up loss
is the mean per-sample loss of each group over the 4 warm-up epochs; ``AUROC''
is the ability of the per-sample warm-up loss to identify corrupted images
(all images / within $g_b$); ``clean@30\%'' is the fraction of clean images
among the 30\% lowest-loss images (clean base rate in parentheses).
$\Delta_{\text{true}}$ is reported for the selected reference and, as a
counterfactual, when inference is conditioned on the other group.}
\label{tab:app_autoscd}
\setlength{\tabcolsep}{4pt}
\begin{tabularx}{\textwidth}{@{} X l c c c c c c @{}}
\toprule
\textbf{Setting} & \textbf{Biased group} & \textbf{Loss} $g_0$ / $g_1$ & \textbf{Selected} & \textbf{AUROC} & \textbf{clean@30\%} & $\boldsymbol{\Delta}_{\text{true}}$ & $\boldsymbol{\Delta}_{\text{true}}$ \textbf{(other)} \\
\midrule
Erosion 50\%, female (majority) & $g_0$ & 0.161 / 0.072 & $g_1$ $\checkmark$ & 0.84 / 0.67 & 1.00 (0.68) & 6.0 & 9.5 \\
Erosion 50\%, male (minority)   & $g_1$ & 0.079 / 0.161 & $g_0$ $\checkmark$ & 0.91 / 0.69 & 1.00 (0.82) & 4.5 & 9.8 \\
Erosion 25\%                    & $g_0$ & 0.139 / 0.071 & $g_1$ $\checkmark$ & 0.99 / 0.99 & 1.00 (0.84) & 1.1 & 1.9 \\
Erosion 100\%                   & $g_0$ & 0.092 / 0.072 & $g_1$ $\checkmark$ & 0.78 / --   & 0.67 (0.37) & 4.8 & 4.5 \\
Dilation 50\%                   & $g_0$ & 0.106 / 0.074 & $g_1$ $\checkmark$ & 0.66 / 0.45 & 0.88 (0.68) & 2.4 & 3.6 \\
HBD 50\%                        & $g_0$ & 0.099 / 0.070 & $g_1$ $\checkmark$ & 0.91 / 0.88 & 0.99 (0.68) & 0.8 & 0.9 \\
Annotator-level (mixed)         & $g_0$ & 0.157 / 0.141 & $g_1$ $\checkmark$ & 0.54 / 0.27 & 0.30 (0.23) & 6.7 & 6.0 \\
Group-independent 50\%          & none  & 0.160 / 0.159 & $g_1$ (undefined) & 0.67 / 0.67 & 0.68 (0.50) & $-2.3$ & $-2.4$ \\
No bias ($\beta=0$)             & none  & 0.077 / 0.069 & $g_1$ (harmless)  & --          & --          & 0.5    & 0.4 \\
\bottomrule
\end{tabularx}
\end{table*}

\end{document}